\documentclass{article}

\usepackage{microtype}
\usepackage{graphicx}
\usepackage{subfigure}
\usepackage{booktabs} 

\usepackage{multirow}
\usepackage{colortbl}

\usepackage{hyperref}

\usepackage[accepted]{icml2025}

\usepackage{amsmath}
\usepackage{amssymb}
\usepackage{mathtools}
\usepackage{amsthm}

\usepackage[capitalize,noabbrev]{cleveref}

\theoremstyle{plain}

\theoremstyle{definition}

\theoremstyle{remark}

\usepackage[textsize=tiny]{todonotes}

\icmltitlerunning{Configuring ICL Demonstrations for Unleashing MLLMs' Sentimental Perception Capability}

\begin{document}

\twocolumn[
\icmltitle{An Empirical Study on Configuring In-Context Learning Demonstrations for Unleashing MLLMs' Sentimental Perception Capability}



\icmlsetsymbol{equal}{*}

\begin{icmlauthorlist}
\icmlauthor{Daiqing Wu}{iie,cas}
\icmlauthor{Dongbao Yang}{iie}
\icmlauthor{Sicheng Zhao}{tsing}
\icmlauthor{Can Ma}{iie}
\icmlauthor{Yu Zhou}{nankai}
\end{icmlauthorlist}

\icmlaffiliation{iie}{Institute of Information Engineering, Chineses Academy of Sciences}
\icmlaffiliation{nankai}{VCIP \& TMCC \& DISSec, College of Computer Science, Nankai University}
\icmlaffiliation{tsing}{Tsinghua University}
\icmlaffiliation{cas}{School of Cyber Security, University of Chinese Academy of Sciences}

\icmlcorrespondingauthor{Dongbao Yang}{yangdongbao@iie.ac.cn}
\icmlcorrespondingauthor{Yu Zhou}{yzhou@nankai.edu.cn}

\icmlkeywords{Machine Learning, ICML}

\vskip 0.3in
]



\printAffiliationsAndNotice{} 

\begin{abstract}
The advancements in Multimodal Large Language Models (MLLMs) have enabled various multimodal tasks to be addressed under a zero-shot paradigm. This paradigm sidesteps the cost of model fine-tuning, emerging as a dominant trend in practical application. Nevertheless, Multimodal Sentiment Analysis (MSA), a pivotal challenge in the quest for general artificial intelligence, fails to accommodate this convenience. The zero-shot paradigm exhibits undesirable performance on MSA, casting doubt on whether MLLMs can perceive sentiments as competent as supervised models. By extending the zero-shot paradigm to In-Context Learning (ICL) and conducting an in-depth study on configuring demonstrations, we validate that MLLMs indeed possess such capability. Specifically, three key factors that cover demonstrations' retrieval, presentation, and distribution are comprehensively investigated and optimized. A sentimental predictive bias inherent in MLLMs is also discovered and later effectively counteracted. By complementing each other, the devised strategies for three factors result in average accuracy improvements of 15.9\% on six MSA datasets against the zero-shot paradigm and 11.2\% against the random ICL baseline. 
\end{abstract}

\section{Introduction}
Equipping models with emotional intelligence has been a fascinating yet vital challenge over the past few decades \cite{tpami2022review, cvpr2023ler}. Studies on various facets of emotions and sentiments in numerous domains have flourished. Among them, Multimodal Sentiment Analysis (MSA) aims to classify the sentiment polarity embedded in multimodal data. As corroborated by empirical \cite{emnlp2017tfn} and theoretical \cite{nips2021m>u} evidence, the synergy between modalities facilitates more comprehensive modeling of sentiment clues compared to unimodal data. This superiority has led to growing interest in MSA from academia and industry \cite{dmkd2018survey1, kis2019survey2}.

\begin{figure}
    \centering
    \includegraphics[width=1\linewidth]{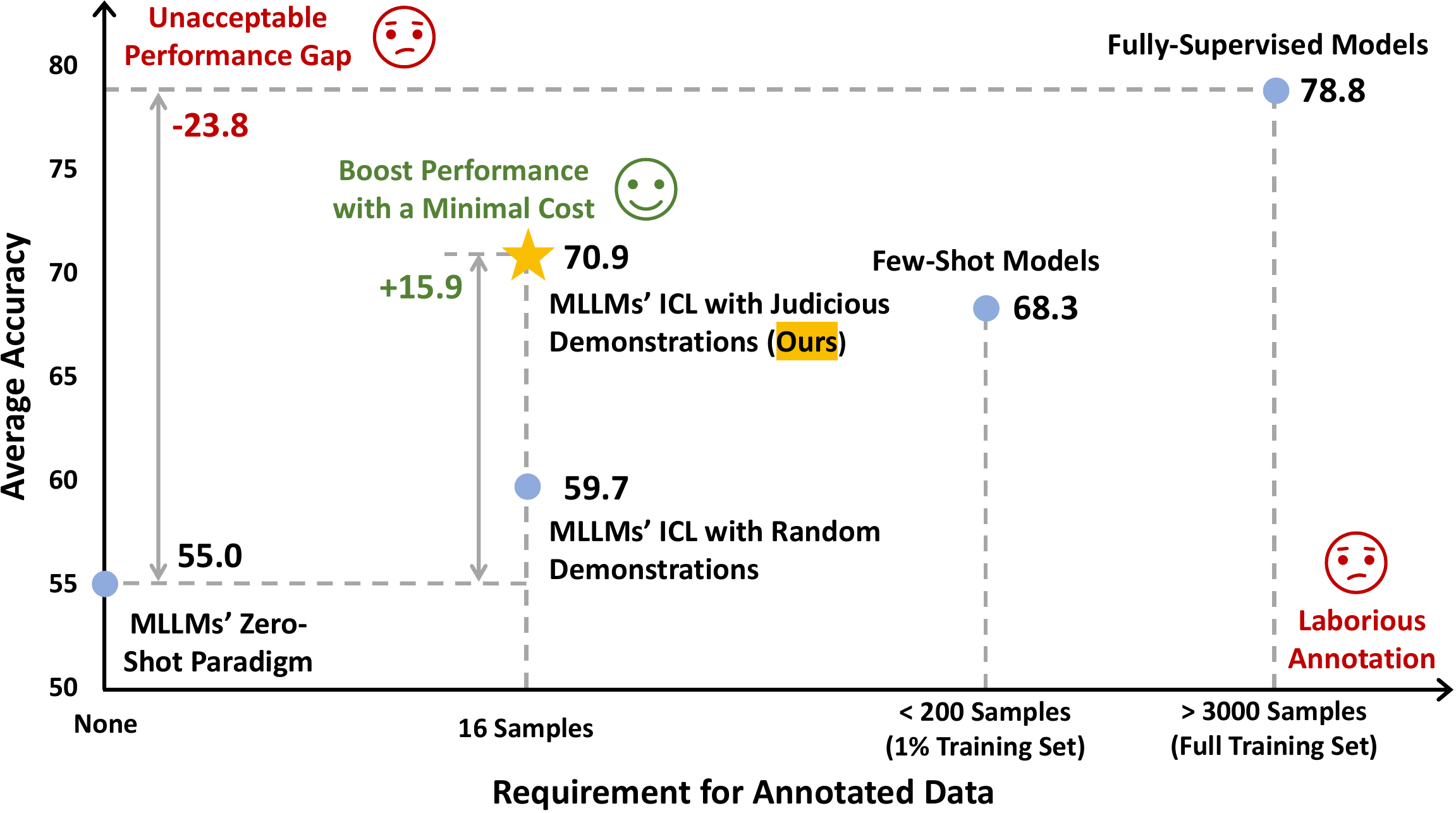}
    \vskip -0.15in
    \caption{Comparison of fully-supervised models, few-shot models, and MLLMs based on average accuracy and annotated data requirement across six MSA datasets. The MLLMs' zero-shot paradigm, although avoiding the laborious annotation, exhibits a substantial performance gap compared to fully-supervised models. With proper demonstration configuration, this gap can be notably narrowed by In-Context Learning (ICL).}
    \vskip -0.15in
    \label{fig1}
\end{figure}

\begin{figure*}
    \centering
    \includegraphics[width=1\linewidth]{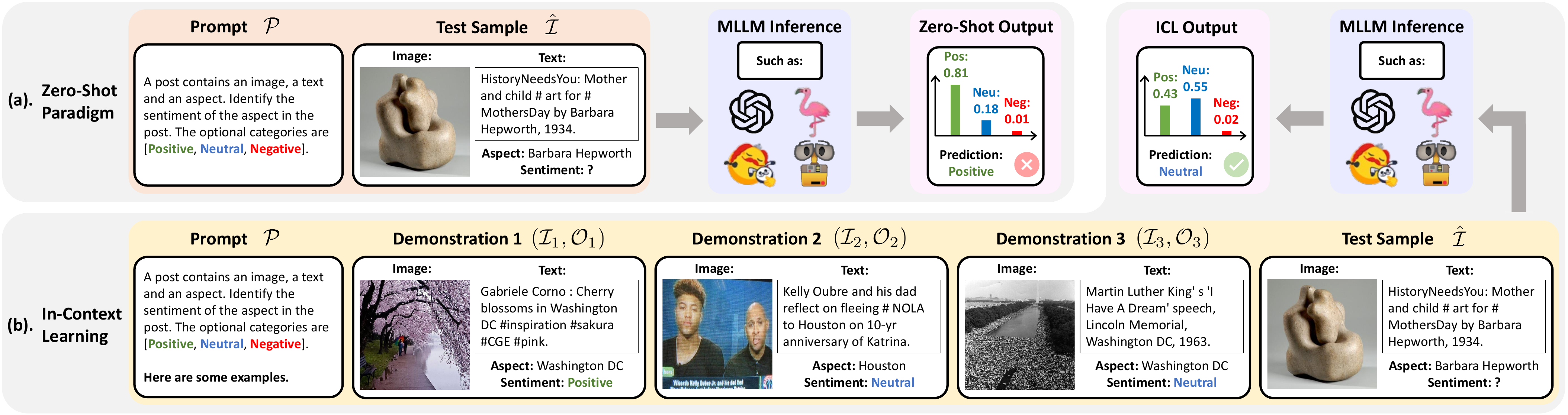}
    \vskip -0.17in
    \caption{Comparison between MLLMs' zero-shot paradigm and ICL. In addition to the test sample, ICL sequences three demonstrations with inputs and corresponding outputs, facilitating more precise sentiment predictions for MLLMs. }
    \vskip -0.15in
    \label{fig2}
\end{figure*}

In this paper, we focus on MSA with the input form of image-text posts, whose number grows exponentially thanks to the prosperity of social media. The current research landscape can be broadly grouped into two streams: post-level branch \cite{cikm2017multisentibank}, which identifies the overall sentiment of posts, and aspect-level branch \cite{aaai2019mimn}, which probes the sentiment associated with specific aspect terms within the context of posts. Leveraging elaborate modules and a large volume of annotated data, both streams have attained remarkable achievements. Entering the era of Multimodal Large Language Models (MLLMs), various multimodal tasks can be reformulated as filling textual prompts and properly accomplished by MLLMs in a zero-shot manner \cite{arxiv2023surveyMLLM}. This emerging paradigm avoids time-intensive fine-tuning and task-specific annotation. However, recent surveys \cite{arxiv2023mmbigbench, if2024gpt4vemotion} have revealed that, under the zero-shot paradigm, MLLMs lag behind supervised models by a significant margin on MSA, as also illustrated in \cref{fig1}. Given the success of MLLMs on tasks such as image captioning \cite{icml2023blip2} and visual question answering \cite{icml2023palme}, it is natural to wonder whether the capabilities of MLLMs to perceive sentiment are yet to be fully explored. If not, how can we fully tap into their potential at a minimal cost?

A feasible answer is In-Context Learning (ICL), which extends MLLMs' zero-shot paradigm to a few-shot scenario by sequencing a series of input-output pairs as demonstrations \cite{neurips2020gpt3}. \cref{fig2} illustrates this process with an example. In this manner, ICL showcases the formulation of the task and the mapping between inputs and outputs \cite{acl2023tr&tl}, which has been proven beneficial for MLLMs in both unimodal and multimodal tasks \cite{arxiv2023visualicl, cvpr2024icl}. Despite its success, the efficacy of ICL heavily relies on the retrieval \cite{emnlp2022r-icl}, presentation \cite{cvpr2024icl}, and distribution \cite{acl2023zicl} of demonstrations. Studying the impact of these factors is a prevalent subject in natural language processing, yet there's a notable absence of necessary attention in multimodal settings, particularly for MSA. 

To fill this gap and unleash the potential of MLLMs, we systematically investigate the configuration of ICL demonstrations in MSA. Specifically, we delve deeper into three crucial factors underexplored in current studies. \textbf{(1). The similarity measurement of multimodal data for demonstration retrieval.} The similarity between the test sample and demonstration is positively correlated with the effectiveness of ICL \cite{deelio2022goodicl}. The mainstream measurement method directly aggregates images' and texts' similarity scores \cite{aaai2022rices}. Nevertheless, our experiments reveal that this approach overlooks fine-grained aspect terms and fails to weigh the significance of modalities. In response, we refine and customize similarity measurements specifically tailored for MSA. \textbf{(2). The trade-off between multimodal information presented in the demonstrations.} Owing to the disparity in information density, images and texts each have unique pros and cons under different scenarios. Therefore, image captioning \cite{mm2023multipoint} and text-to-image generation \cite{arxiv2024t2iapp} are commonly employed to convert between the two modalities to furnish supportive information in multimodal tasks. Inspired by this, we explore the effects of modality compositions on ICL, ultimately arriving at the most efficacious modality presentation. \textbf{(3). The biases introduced by sentiment distribution.} It has been observed on various tasks \cite{nips2023icicl, cvpr2024icl, cvprw2024rebut} that MLLMs are prone to be affected by the biases in ICL sequences, a phenomenon we also validate on MSA. Driven by this discovery, we devise various distribution protocols to intentionally incorporate sentiment biases and probe their influences. Comprehensive experiments reveal that appropriate sentiment biases can counterbalance the inherent predictive bias of MLLMs, thereby promoting fair prediction. In summary, our contributions are three-fold:

\vskip -0.15in
\begin{itemize}
    \item Through configuring ICL demonstrations, we unleash the potential of MLLMs on MSA, validating that MLLMs are competent in perceiving sentiment.
    \vskip -0.15in
    \item We investigate and optimize three key factors covering the retrieval, presentation, and distribution of demonstrations in ICL on MSA. During the process, a sentimental predictive bias inherent in MLLMs is discovered and mitigated, facilitating fairness in sentiment prediction.
    \vskip -0.15in
    \item By complementing each other, the ICL strategies tailored for the three factors improve the accuracy of MLLMs by an average of 15.9\% on six MSA datasets compared to the zero-shot paradigm, and 11.2\% compared to the random ICL baseline.
\end{itemize}
\vskip -0.15in

\begin{figure*}
    \centering
    \includegraphics[width=1\linewidth]{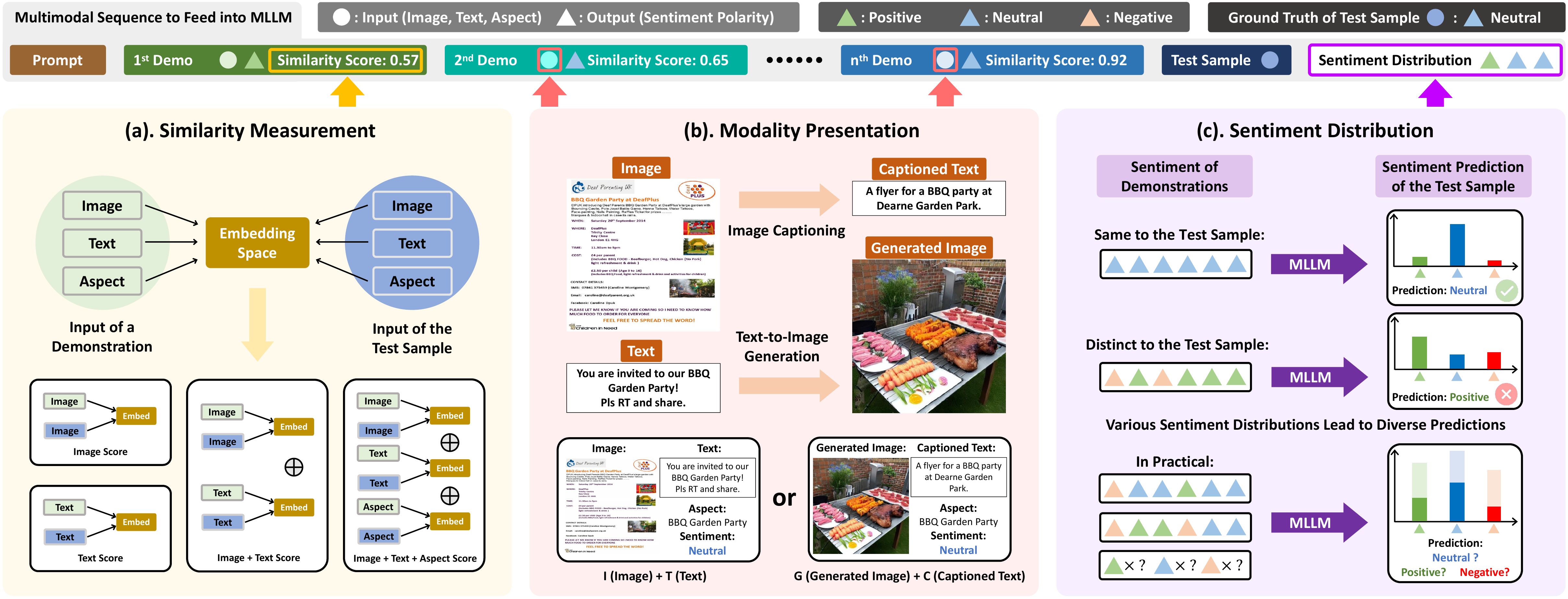}
    \vskip -0.15in
    \caption{Illustration of the three factors to be investigated and optimized, during which we aim to address the following questions. (a). How do we measure the similarity score between multimodal data? (b). How do we decide which modality should be presented in the input? (c). What kind of impact does the sentiment distribution of demonstrations have? }
    \vskip -0.15in
    \label{fig3}
\end{figure*}

\section{Related Works}
\subsection{Multimodal Sentiment Analysis}
With the explosion of multimodal data from social media platforms in recent years, MSA has evolved as a pivotal topic in affective computing \cite{cc2024msareview, arxiv2024msallmsurvey}. Image-text posts \cite{cikm2017multisentibank} and videos \cite{emnlp2017tfn} are two major forms of input in MSA. Our investigation centers on image-text posts, as they are more congruent with the input specifications of MLLMs. Within this domain, the majority of research adheres to a similar methodology, typically initiating with the extraction of unimodal features followed by their fusion for sentiment classification. Over the past few years, notable advancements have been made in both post-level \cite{isi2017hsan, acl2021mgnns, naacl2022clmlf, acl2023mvcn, mm2024drf} and aspect-level \cite{sigir2018comn, acl2022vlp, acl2023aom, aaai2024ebmm} branches,  by refining fusion strategies and learning from thousands of hand-annotated data. However, as highlighted by \citet{mm2022up}, the acquisition of such data is unrealistic in practical settings. Consequently, few-shot MSA has been proposed \cite{mm2022up} and developed \cite{mm2023multipoint}. Although it reduces data requirements, the time-intensive fine-tuning persists as an impediment to the cost-effective application of MSA.

The advent of MLLMs presents a viable alternative via the zero-shot paradigm \cite{if2024gpt4vemotion}, which is subsequently augmented by ICL. By harnessing the general knowledge acquired during pre-training, MLLMs are endowed with the proficiency to address a wide array of downstream tasks, requiring merely a minimal set of annotated samples and no gradient updates. Compared with fully-supervised and few-shot MSA, the ICL of MLLMs accords better with practical applications, holding the potential to evolve into a prevailing trend for future development.


\subsection{In-Context Learning}
The scaling up of model size has empowered Large Language Models (LLMs) to generalize to unseen tasks through analogical learning \cite{arxiv2022iclsurvey}, a capability known as ICL \cite{arxiv2024mllmsurvey}. First identified in GPT-3 \cite{neurips2020gpt3}, ICL has sparked numerous studies on LLMs. Among these, a predominant emphasis has been placed on the impact of demonstration configurations. Through exhaustive explorations, various effective strategies have been proposed from distinct angles \cite{acl2022order, emnlpf2023support, acl2023saicl, acl2023coverls}. Certain ICL strategies of LLMs also demonstrate proficiency in tackling multimodal tasks \cite{aaai2022rices, emnlpf2023ner} by converting them into textual format.

To transfer the ICL capability into MLLMs, Flamingo \cite{neurips2022flamingo} interleaves cross-attention modules that receive visual input into the pre-trained LLMs. Inspired by its success, a series of open-sourced MLLMs armed with ICL capability are developed, such as Open-Flamingo \cite{arxiv2023openflamingo}, IDEFICS \cite{nips2023obelics} and Kosmos \cite{neurips2023kosmos}. Parallel to the case in LLMs, the performance of MLLMs exhibits significant variability in response to diverse demonstrations. A line of studies methodically probes into such effects in image captioning \cite{nips2023icicl} and visual question answering \cite{cvpr2024icl, iclrw2024mmices, neurips2024icl}, uncovering valuable insights of ICL and important properties of MLLMs. Nonetheless, the unique challenges intrinsic to MSA remain inadequately overcome. We aim to fill this blank through a comprehensive empirical exploration.

\section{Configuring ICL Demonstrations}
\subsection{Task Definition}
MSA is a classification task where the target categories are sentiment polarities.
Depending on the classification objective, MSA is grouped into post-level MSA and aspect-level MSA. For post-level MSA, the input comprises an image-text post $(\boldsymbol{i}, \boldsymbol{t})$, and the target is to identify the overall sentiment. For aspect-level MSA, the input includes an additional aspect $\boldsymbol{a}$: $(\boldsymbol{i}, \boldsymbol{t}, \boldsymbol{a})$, and the target shifts to identify the sentiment of the aspect within the context of the post. We wrap the input as $\mathcal{I}$ and the output as $\mathcal{O}$ for uniformity.

Before feeding into the MLLM, the test sample $\hat{\mathcal{I}}$ and $n$ demonstrations $(\mathcal{I}, \mathcal{O})$ are templatized into a multimodal sequence $\mathcal{S} = \{\mathcal{P};(\mathcal{I}_1, \mathcal{O}_1);(\mathcal{I}_2, \mathcal{O}_2);\ldots;(\mathcal{I}_n, \mathcal{O}_n);\hat{\mathcal{I}}\}$, where $\mathcal{P}$ is a textual prompt including the task description and the set of target categories. A 3-shot case is illustrated in \cref{fig2} (b). The output $\hat{\mathcal{O}}$ is subsequently generated by MLLM $\boldsymbol{M}(\cdot)$ predicting the next single token:
\begin{equation}
\label{Task definition}
\hat{\mathcal{O}} =  \mathop{\arg\max}\limits_{\mathcal{T}}\boldsymbol{M}(\mathcal{T}|\mathcal{S}).
\end{equation}
Achieving desirable outcomes through ICL involves the interplay of various factors that each contribute to the process. Among these, the retrieval, presentation, and distribution of demonstrations have been empirically verified to wield significant impact \cite{emnlp2022r-icl, cvpr2024icl, acl2023zicl}. In light of MSA's multimodal and affective nature, we concretize these aspects into three key factors to optimize: similarity measurement, modality presentation, and sentiment distribution. The three factors are briefly illustrated in \cref{fig3} (a), (b), and (c), respectively, and are elaborated on in the subsequent sections.

\subsection{Similarity Measurement}
In demonstration retrieval, prior research has reached two primary agreements: demonstrations bearing greater similarity to the test sample are more beneficial for the reasoning process of MLLMs \cite{deelio2022goodicl}, and demonstrations should be ordered from lowest to highest similarity in the ICL sequence \cite{acl2022order}. Given adherence to these two fundamental principles, measuring the similarity between multimodal data emerges as the crux of the matter.

Considering the input of a demonstration $(\boldsymbol{i}, \boldsymbol{t}, \boldsymbol{a})$ and the test sample $(\boldsymbol{\hat{i}}, \boldsymbol{\hat{t}}, \boldsymbol{\hat{a}})$, there are currently three predominant strategies to measure similarity. Here we discuss the scenario of aspect-level MSA due to its generalizability.

\noindent
\textbf{Image Based (I)} and \textbf{Text Based (T)} strategies. Under unimodal settings, the similarity score $\mathcal{K}$ can be straightforwardly obtained by computing the cosine similarity in the embedding space of a pre-trained encoder $\mathcal{E}$. Indexing by images, this process can be described as: 
\begin{equation}
\label{similarity calaulation}
\mathcal{K}_{I} = \frac{\mathcal{E}(\boldsymbol{i})\odot \mathcal{E}(\boldsymbol{\hat{i}})}{||\mathcal{E}(\boldsymbol{i})||\cdot||\mathcal{E}(\boldsymbol{\hat{i}})||}.
\end{equation}
For simplicity, we abbreviate it as $\mathcal{K}_{I} = C(\boldsymbol{i}, \boldsymbol{\hat{i}})$. Likewise, with text indexing, it would be $\mathcal{K}_{T} = C(\boldsymbol{t}, \boldsymbol{\hat{t}})$.

\noindent
\textbf{Image-Text Based (IT)} strategy. It extends the unimodal strategies to a multimodal form by aggregating the unimodal similarity scores: $\mathcal{K}_{IT} = \mathcal{K}_{I} + \mathcal{K}_{T}$. This strategy is also referred to as RICES \cite{aaai2022rices}.

Despite their proven efficacy across various tasks, these strategies fall short in two critical areas when applied to MSA. In particular, they fail to consider the aspect-specific relevance and overlook the unequal significance of modalities. To probe the impact of these shortcomings and optimize accordingly, we further devise six strategies in \cref{tab1}. Among these, the \textbf{Aspect Based (A)} strategy is an unimodal measurement indexed by aspects. The \textbf{Image-Aspect Based (IA)}, \textbf{Text-Aspect Based (TA)}, and \textbf{Image-Text-Aspect Based (ITA)} strategies are the aspect-inclusive versions of the \textbf{I}, \textbf{T}, and \textbf{IT} strategies, respectively. The \textbf{Weighted Image-Text Based (WIT)} and \textbf{Weighted Image-Text-Aspect Based (WITA)} strategies expand upon the \textbf{IT} and \textbf{ITA} strategies by incorporating weights $\boldsymbol{\alpha}, \boldsymbol{\beta}, \boldsymbol{\gamma}$ to modulate the impact of each modality.

\begin{table}[t]
\vskip -0.07in
\centering
\caption{Formulations of six devised strategies for measuring similarity between multimodal data.}
\vskip 0.05in
\resizebox{0.8\linewidth}{!}{
\begin{tabular}{l|l}
\toprule
Strategy & Formulation \\
\midrule
A        & $\mathcal{K}_{A} = C(\boldsymbol{a}, \boldsymbol{\hat{a}})$  \\
IA       & $\mathcal{K}_{IA} = \mathcal{K}_{I} + \mathcal{K}_{A}$  \\
TA       & $\mathcal{K}_{TA} = \mathcal{K}_{T} + \mathcal{K}_{A}$  \\
ITA      & $\mathcal{K}_{ITA} = \mathcal{K}_{I} + \mathcal{K}_{T} + \mathcal{K}_{A}$  \\
WIT      & $\mathcal{K}_{WIT} = \boldsymbol{\alpha} \cdot \mathcal{K}_{I} + \boldsymbol{\beta} \cdot \mathcal{K}_{T}$            \\
WITA     & $\mathcal{K}_{WITA} = \boldsymbol{\alpha} \cdot \mathcal{K}_{I} + \boldsymbol{\beta} \cdot \mathcal{K}_{T} + \boldsymbol{\gamma} \cdot \mathcal{K}_{A}$  \\
\bottomrule
\end{tabular}}
\vskip -0.10in
\label{tab1}
\end{table}

\subsection{Modality Presentation}
The density of information embedded in images and texts naturally differs \cite{cvpr2023fdt, mm2024pacl}. Images convey information in a broadly ranged yet abstract manner, whereas text is typically more precise and concise. By converting images to text through image captioning, salient objects are emphasized and the irrelevant fine-grained details are omitted \cite{icml2015sat, nips2023icicl}. Conversely, transforming texts into images through text-to-image generation can supplement the background and context beyond the description of texts. The extra information is conducive to the elicitation of previously latent emotions \cite{cvpr2017emotic}. As exemplified in \cref{fig3} (b), the captioned text summarizes the flyer's contents, and the generated image portrays the unmentioned details of the barbecue. Recognizing these benefits, various multimodal studies leverage these techniques to derive auxiliary modalities to promote performance. 

This drives us to investigate whether such success can be replicated in multimodal ICL, where the form in which information is presented within the ICL sequence is of critical importance. Specifically, we obtain the captioned texts from the MLLM itself, and the generated images from the diffusion model \cite{cvpr2022diffusion}. Subsequently, we reconstruct the inputs of demonstrations and the test sample with combinations of original and auxiliary modalities. An input form combining the captioned text and generated image is depicted at the bottom of \cref{fig3} (b), and a broader range of combinations is investigated in the experiments.

\subsection{Sentiment Distribution}
Under ICL, MLLMs have been shown to manifest a short-cut effect \cite{acl2023zicl}. Upon encountering a test sample, MLLMs tend to duplicate the output from one of the demonstrations, and this tendency intensifies when multiple demonstrations share identical outputs. This exposes MLLMs' susceptibility to biases presented within the ICL sequence, where divergent biases can lead to markedly different predictions. 

Regarding MSA, its classification nature and limited categories inevitably result in certain sentiment biases within the ICL sequence, reflected in the sentiment distribution of demonstrations. To analyze the potential impact of these biases and utilize them to optimize the ICL configuration, we formulate five distribution protocols that regulate the sentiment distribution of demonstrations, thereby injecting distinct sentiment biases. The first two protocols serve to highlight the spectrum of the impact of biases. \textbf{(1). Ideal} protocol: every demonstration aligns with the test sample's sentiment. \textbf{(2). Contrary to Ideal} protocol: every demonstration differs from the test sample's sentiment. Utilizing the ground truth from test samples, they introduce extreme biases into the ICL sequence to establish the theoretical ceiling and floor. In contrast, the following protocols are devised for practical ICL configuration. \textbf{(3). Unlimited} protocol: demonstrations are retrieved without distribution restriction. \textbf{(4). Category Balanced} protocol: each sentiment class carries an identical number of demonstrations. \textbf{(5). Identical to Support Set} protocol: the sentiment distribution of demonstrations mirrors that of the support set.

\begin{table}[t]
\vskip -0.15in
\centering
\caption{Average accuracy across 4,8,16-shot demonstrations retrieved based on varying similarity measurements. R strategy represents the random retrieval. }
\vskip 0.05in
\small
\begin{tabular}{lcccc}
\toprule
\multirow{2}{*}{Strategy} & \multicolumn{2}{c}{Post-Level} & \multicolumn{2}{c}{Aspect-Level} \\
\cmidrule(l){2-3} \cmidrule(l){4-5} 
        & MVSA-S           & MVSA-M          & Twitter-15      & Twitter-17     \\
\midrule
R      & 49.2             & 60.8            & 57.4            & 56.4           \\
\midrule
I      & \textbf{56.5}    & 64.9            & 59.1            & 56.7           \\
T      & 56.0             & \textbf{66.2}   & 58.7            & 57.0           \\
IT     & 55.7             & 66.2            & 61.4            & 57.6           \\
\midrule
A    & -                &  -              & \textbf{61.3}   & 57.4           \\
IA      & -                &  -              & 59.6            & 58.0           \\
TA     & -                &  -              & 60.9            & 57.3           \\
ITA     & -                &  -              & 61.0            & \textbf{58.1}  \\
\bottomrule
\end{tabular}
\vskip -0.15in
\label{tab2}
\end{table}

\section{Experiments}
\subsection{Datasets and Implementation Details}
In exploratory experiments, IDEFICS-9B \cite{nips2023obelics} is selected as a representative MLLM \cite{iclr2024beyond}. Datasets MVSA-S, MVSA-M \cite{mmm2016mvsa} are utilized for post-level MSA, and Twitter-15 \cite{aaai2018t15}, Twitter-17 \cite{acl2018t17} are chosen for aspect-level MSA. For each dataset, demonstrations are retrieved from the support set sampled as \citet{mm2023multipoint}, which accounts for 1\% of the training set. MLLMs' performances are evaluated on the original test set, with accuracy serving as the primary metric. The three factors are investigated separately, and the optimal strategies are subsequently combined for a comprehensive assessment. To validate the generalizability, additional experiments are carried out using Open-Flamingo2-9B \cite{arxiv2023openflamingo} and on two other MSA datasets TumEmo \cite{tmm2021tumemo} and MASAD \cite{nc2021masad}.

Due to practical considerations, we choose the above model scale, which also aligns with the primary research scope of \citet{nips2023icicl, cvpr2024icl}.
By default, we adopt the \textbf{IT} strategy for similarity measurement, compose the input by the original image and text, and put no constraints on the distribution. Variations are introduced only to the pertinent settings when investigating a specific factor.

\subsection{Results and Analysis}

\begin{figure}[t]
    \centering
    \includegraphics[width=1\linewidth]{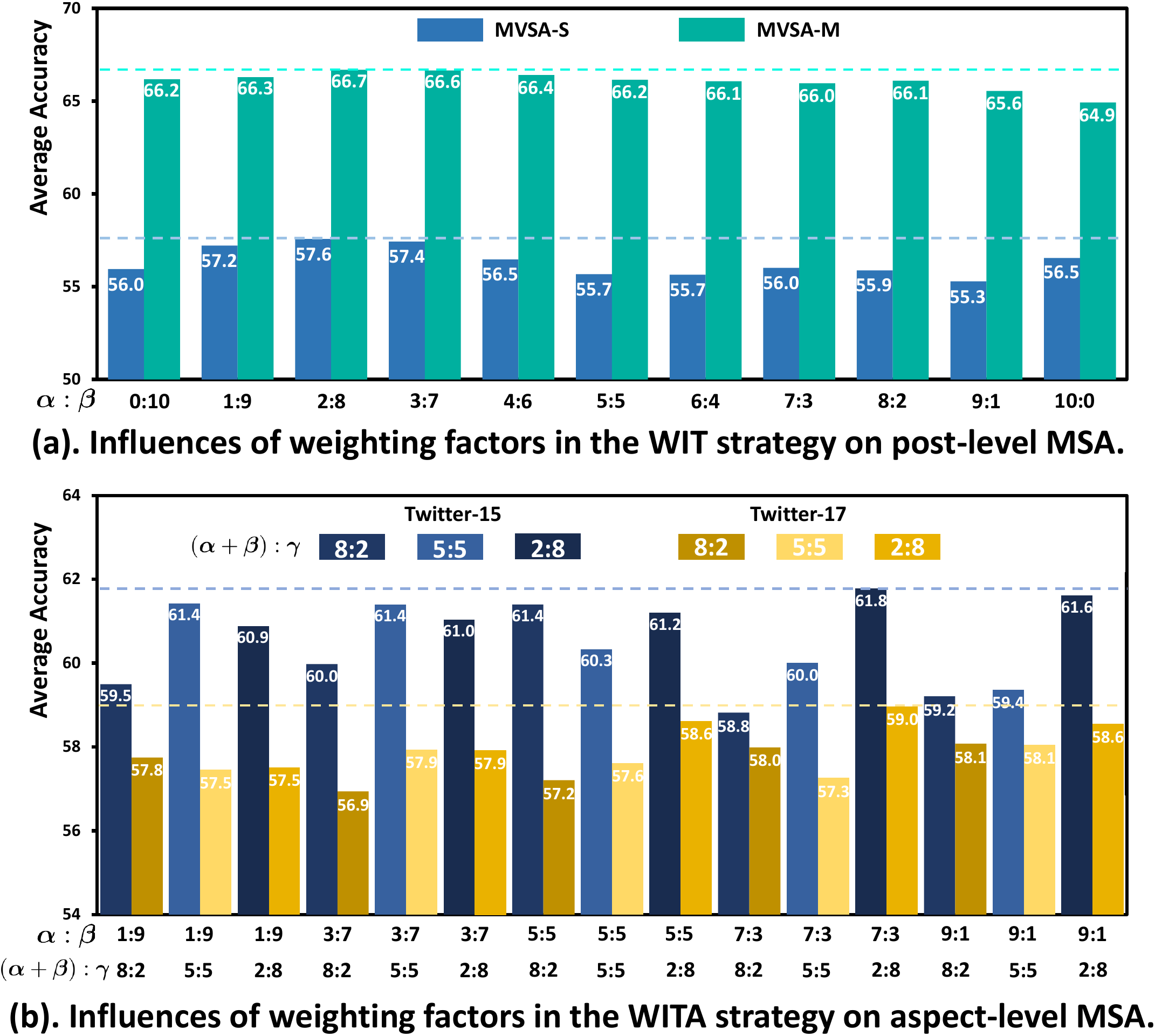}
    \vskip -0.10in
    \caption{Average accuracy across 4,8,16-shot demonstrations retrieved based on the WIT and WITA strategies. }
    \label{fig4}
    \vskip -0.10in
\end{figure}

\subsubsection{Similarity Measurement}
\cref{tab2} contrasts the impact of similarity measurement. Here we employ CLIP \cite{icml2021clip} as the pre-trained encoder for both images and texts following previous studies \cite{aaai2022rices, cvpr2024icl}. Observably, all measurements surpass random retrieval, revealing their validity in reflecting the relationships among multimodal samples. Moreover, in aspect-level MSA, the consistent improvement achieved by factoring in aspect similarity underscores the aspect's significance. Counterintuitively, the \textbf{IT} or \textbf{ITA} strategies do not always yield optimal results, despite considering the most comprehensive information. Explanations for this necessitate an analysis in conjunction with the results of \textbf{WIT} and \textbf{WITA} strategies. As evidenced in \cref{fig4}, the performance of these strategies fluctuates in response to alterations in the weighting factors. This emphasizes the criticality of maintaining a proper balance among similarity components, which the \textbf{IT} and \textbf{ITA} strategies fail to achieve, thereby leading to their deficiency.

Delving into the impact of weighting factors, we uncover that prioritizing textual similarity ($\boldsymbol{\alpha} < \boldsymbol{\beta}$) yields a pronounced benefit for the \textbf{WIT} strategy in post-level MSA. This benefit peaks at $\boldsymbol{\alpha}:\boldsymbol{\beta}=2:8$, where the \textbf{WIT} strategy achieves an accuracy of 57.6 on MVSA-S and 66.7 on MVSA-M, outperforming all competing strategies. In aspect-level MSA, as illustrated in \cref{fig4} (b), the \textbf{WITA} strategy should, instead, accord priority to the similarity of aspects. Interestingly, under this priority, the relationship between image and text weights is reversed. We attribute these to the fact that aspects are most directly correlated with sentiment prediction, and the similarity among aspects tends to eclipse the textual similarity, given that these aspects typically originate from the texts. When $\boldsymbol{\alpha}:\boldsymbol{\beta}=7:3$ and $(\boldsymbol{\alpha}+\boldsymbol{\beta}):\boldsymbol{\gamma}=2:8$, the \textbf{WITA} strategy achieves peak accuracy of 61.8 on Twitter-15 and 59.0 on Twitter-17, making it the optimal strategy for aspect-level MSA.

\begin{table}[t]
\vskip -0.12in
\centering
\small
\caption{Average accuracy over 4,8,16-shot settings with the inputs composed of different modalities. ``I'' abbreviated for Image, T--Text, C--Captioned Text, G--Generated Image.}
\vskip 0.05in
\begin{tabular}{lccccc}
\toprule

\makebox[0.10\linewidth][l]{\multirow{2}{*}{Modality}} & \multicolumn{2}{c}{Post-Level} & \multicolumn{2}{c}{Aspect-Level} & \makebox[0.08\linewidth][c]{\multirow{2}{*}{Mean}}\\
\cmidrule(l){2-3} \cmidrule(l){4-5} 
        & \makebox[0.12\linewidth][c]{MVSA-S}     & \makebox[0.12\linewidth][c]{MVSA-M}          & \makebox[0.12\linewidth][c]{Twitter-15}      & \makebox[0.12\linewidth][c]{Twitter-17}     \\
\midrule
I      & 51.7             & 56.7            & 57.4            & 53.4       &54.8    \\
C      & 43.9             & 49.5            & 56.2            & 52.1       &50.4    \\
I,C    & 56.9             & 59.3            & 54.3            & 50.1       &55.3    \\
\midrule
T      & 46.3             & 56.0            & 61.7            & \textbf{59.0}       &55.8    \\
G      & 46.8             & 55.5            & 54.8            & 55.2       &53.1    \\
T,G    & 47.7             & 58.1            & 54.0            & 54.8       &53.6    \\
\midrule
I,T    & \textbf{55.7}   & \textbf{66.2}   & 61.4            & 57.6 &\textbf{60.2} \\
I,G    & 54.5             & 58.8            & 54.5            & 53.8       &55.4    \\
C,T    & 49.7             & 60.3            & \textbf{62.6}   & 56.7       &57.3    \\
C,G    & 47.5             & 53.7            & 53.1            & 53.1       &51.8    \\
\midrule
I,C,T  & 55.3             & 64.4            & 60.3            & 55.6       &58.9    \\
I,T,G  & 50.8             & 62.0            & 59.7            & 54.8       &56.8    \\
C,T,G  & 47.4             & 58.2            & 56.5            & 54.1       &54.1    \\
I,C,G  & 50.3             & 57.0            & 52.9            & 52.1       &53.1    \\
I,C,T,G& 49.2             & 61.4            & 58.7            & 53.1       &55.6    \\
\bottomrule
\end{tabular}
\vskip -0.15in
\label{tab3}
\end{table}

\begin{figure}[t]
    \centering
    \includegraphics[width=1\linewidth]{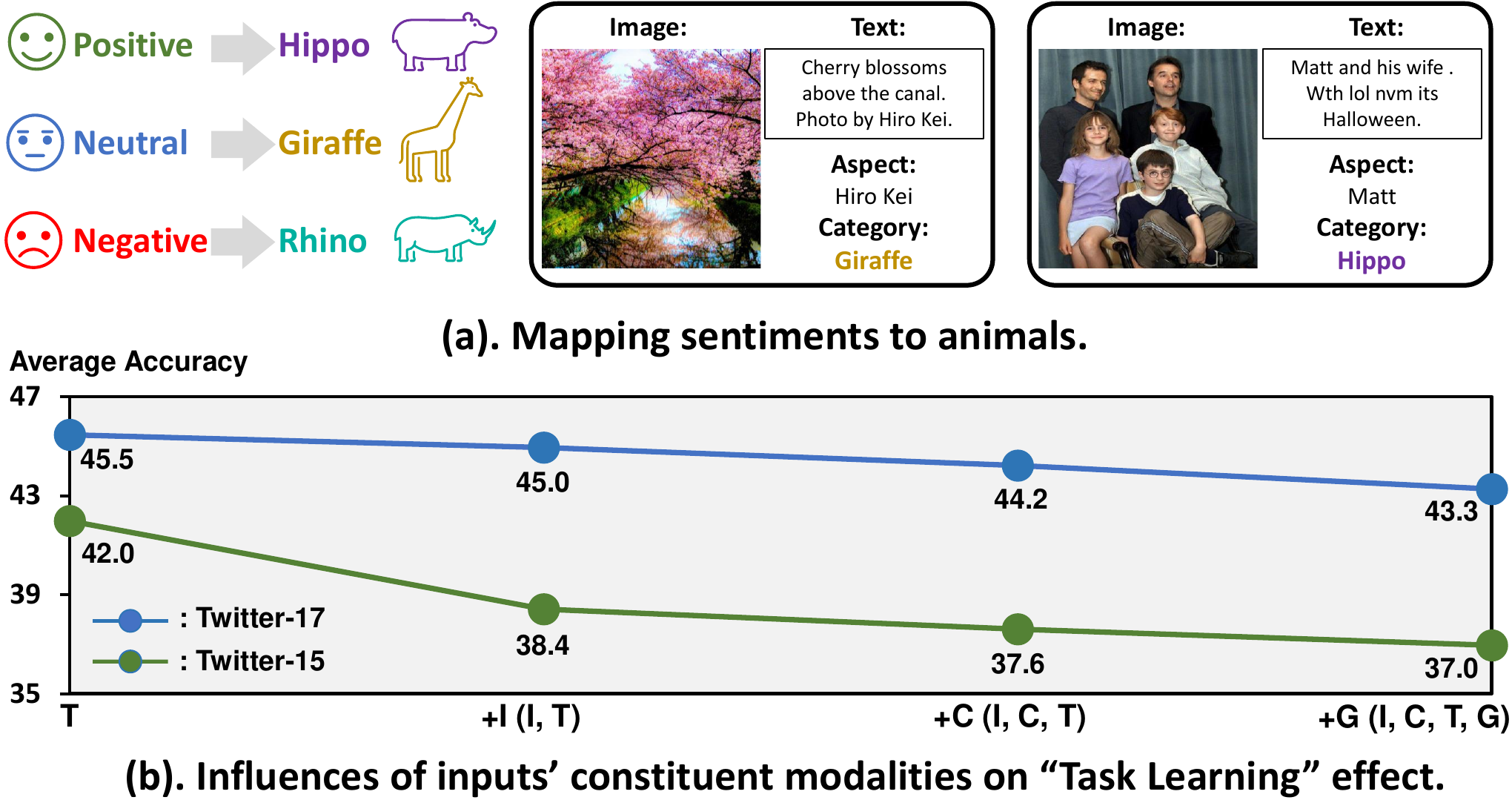}
    \vskip -0.10in
    \caption{Evaluation of ICL's ``Task Learning'' effect by progressively incorporating modalities into the inputs. }
    \vskip -0.10in
    \label{fig5}
\end{figure}

\subsubsection{Modality Presentation}
\cref{tab3} compares inputs composed of various modalities. When input is confined to single-modal information, presenting the text modality generally leads to superior outcomes, reaffirming the significance of texts in MSA. As more modalities are incorporated, the MLLM can benefit from processing multimodal information, thereby enhancing sentiment prediction in most cases. Noticeably, substituting original modalities with auxiliary ones (e.g. from (I, T) to (C, G)) leads to universal performance degradation. This implies that the potential benefits of modality conversions, which involve highlighting salient objects or providing richer context, are outweighed by the inherent drawbacks. These drawbacks encompass information loss and the generation of extraneous noise during the conversion process. Strangely, augmenting original modalities with auxiliary ones (e.g. from (I, T) to (I, C, T)) still results in a performance drop, despite the pure addition of information without any loss. Drawing from insights in \citet{acl2023tr&tl}, we associate this with the impairment of the ``Task Learning'' effect in ICL. Specifically, \citet{acl2023tr&tl} decompose ICL’s role into ``Task Recognition'' and ``Task Learning''. The former prompts the task format for MLLMs to apply their prior knowledge, and the latter aids MLLMs in building mapping relationships between inputs and outputs. Incorporating additional modalities complicates the inputs, making it more challenging for the MLLM to learn these relationships accurately.

\begin{figure*}[t]
    \centering
    \includegraphics[width=1\linewidth]{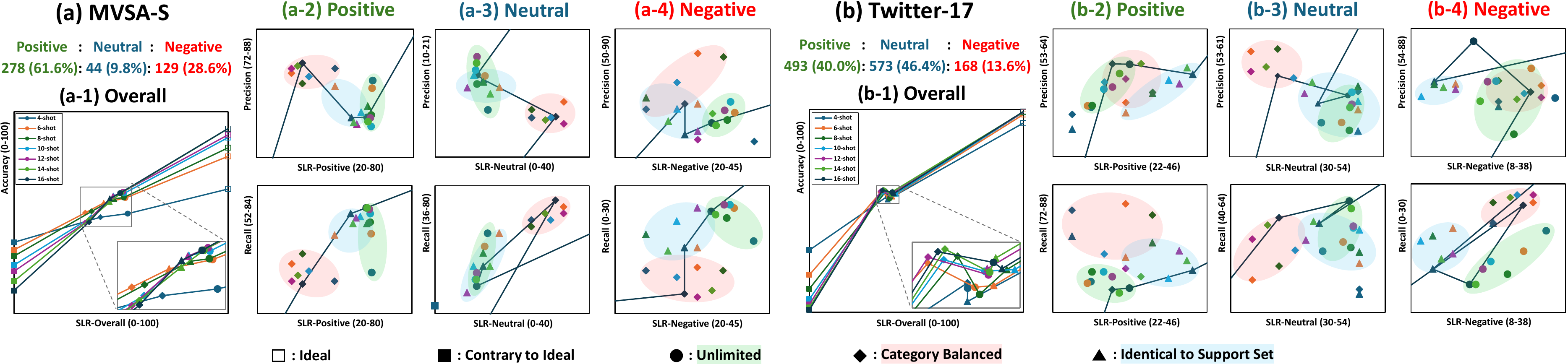}
    \vskip -0.10in
    \caption{Impact of distribution protocols on MLLM's sentiment predictions. In addition to overall accuracy, precision and recall for each category are studied in 2-16 shot settings. The two selected datasets possess distinctive test set distributions, which are detailed above the accuracy charts. In (a-2,3,4) and (b-2,3,4), the results of the same protocol under various shots are grouped into clusters of three colors, and the results of different protocols under 16-shot are connected in a fixed order.}
    \vskip -0.10in
    \label{fig7}
\end{figure*}

To validate this, we design the experiments in \cref{fig5} (a) to quantitatively evaluate the ``Task Learning'' effect. Before feeding into the MLLM, each sentiment is substituted with an animal unrelated to the input semantics according to a pre-defined mapping. This approach ensures that the MLLM can no longer rely on pre-trained prior knowledge but is instead compelled to learn the input-output mapping solely from ICL. As shown in \cref{fig5} (b), the ``Task Learning'' effect diminishes as additional modalities are incorporated into the input, supporting our explanation. To this end, employing original images and text to form inputs has empirically proven optimal, as it attains the finest equilibrium between information enrichment and input complexity.

\begin{figure}[t]
    \centering
    \includegraphics[width=1\linewidth]{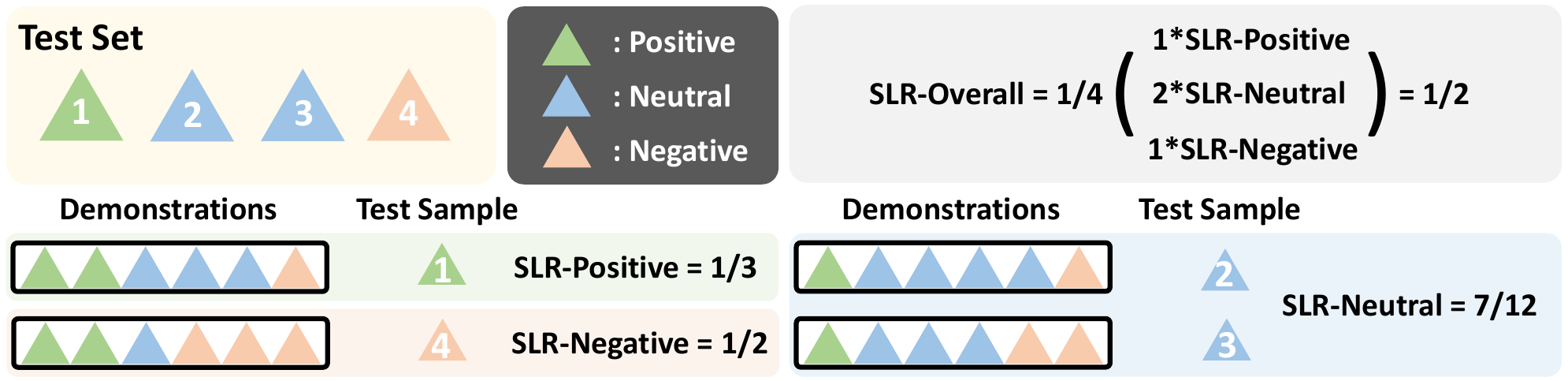}
    \vskip -0.10in
    \caption{Illustration of SLR under 6-shot setting. Taking SLR-Neutral as an example, the ICL sequences of the two Neutral test samples contain 4 and 3 Neutral demonstrations respectively, thus SLR-Neutral is computed as $(3/6+4/6)/2=7/12$.}
    \label{fig6}
    \vskip -0.15in
\end{figure}

\subsubsection{Sentiment Distribution}
When assessing the impact of sentiment biases within the ICL sequence, it is intuitive to assume that the more demonstrations sharing the same sentiment as the test sample, the more likely the MLLM will make the correct prediction. Indeed, this is one of the primary mechanisms by which distribution protocols affect the performance of MLLMs. To quantify this, we introduce a metric termed ``Same Label Rate (SLR)''. First, the proportion of demonstrations with identical sentiments to each test sample is calculated. Then, these proportions are averaged across a collection of test samples. Depending on whether the collection comprises the entire test set or all test samples of a specific category, SLR can be calculated for all test samples (SLR-Overall) or the corresponding sentiment (e.g., SLR-Positive). \cref{fig6} provides an example illustrating SLR. Utilizing SLR, \cref{fig7} evaluates the nuanced impact of distribution protocols.


On MVSA-S, when the number of shots is fixed, overall accuracy exhibits a stable positive correlation with SLR-Overall (\cref{fig7} (a-1)). Under the extreme sentiment biases imposed by the \textbf{Ideal} and \textbf{Contrary to Ideal} protocols, MLLM's sentiment prediction tends to adapt correspondingly, with this trend becoming more pronounced as the number of shots increases. It validates the vulnerability of MLLM to sentiment biases within the ICL sequence. Inspecting each sentiment category (\cref{fig7} (a-2,3,4)), MLLM performs notably well on the positive samples, while it shows considerably diminished precision on neutral samples and recall on negative samples. This reveals a potential predictive bias in the MLLM, which favors positive and neutral over negative. Among three practical protocols, the \textbf{Unlimited} protocol stands out with superior performance, particularly in positive and negative samples. Compared to the \textbf{Contrary to Ideal} protocol, it attains a higher SLR in these two categories. This indicates that similarity-based retrieval improves SLR upon the original distribution in the support set. Though resulting in the least favorable performance, the \textbf{Category Balanced} protocol exhibits an intriguing phenomenon. It achieves the highest SLR in the rarest category (Neutral) while obtaining the lowest SLR in a more prevalent category (Negative), leading the model to classify samples of the latter as the former.

\begin{figure}[t]
    \vskip -0.20in
    \centering
    \includegraphics[width=0.7\linewidth]{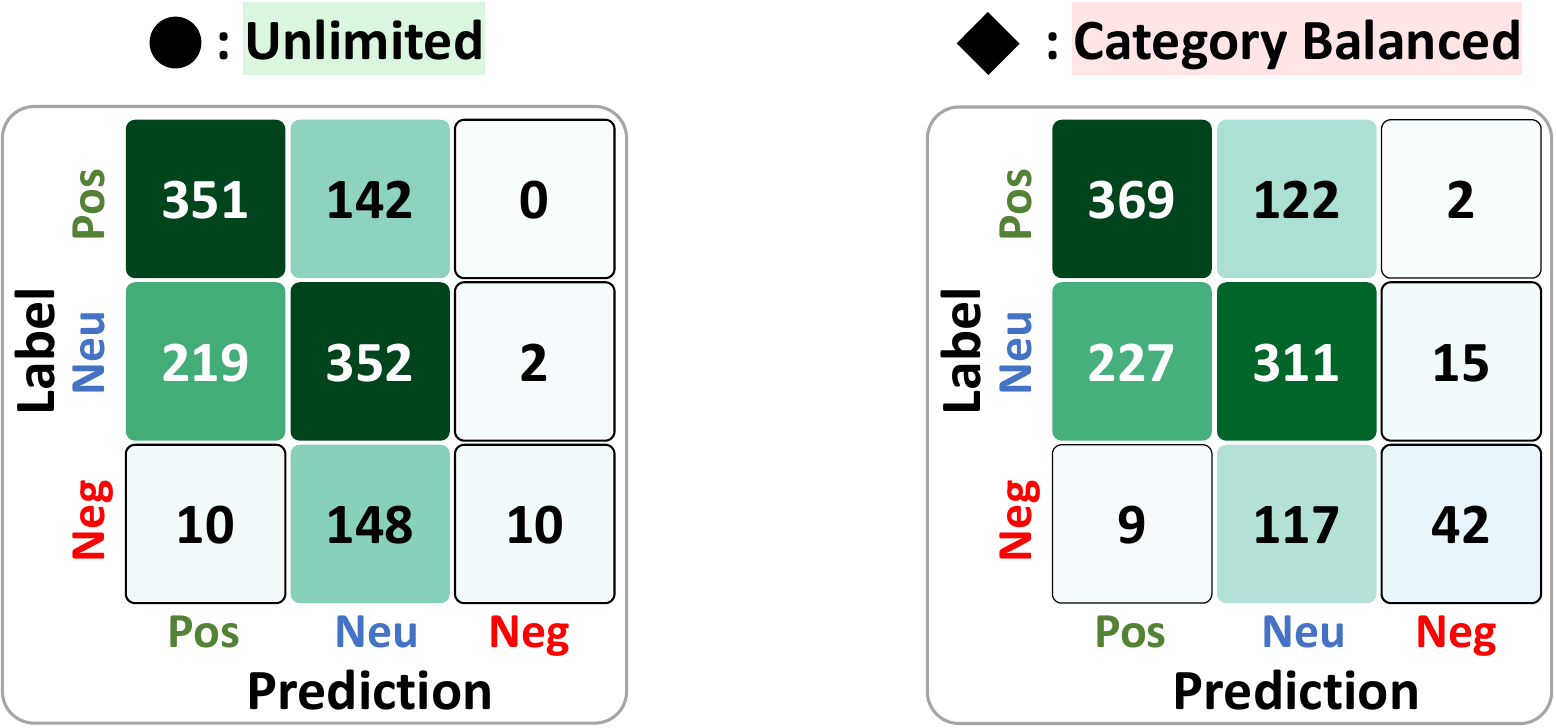}
    \vskip -0.10in
    \caption{Confusion matrices of IDEFICS under the \textbf{Unlimited} and \textbf{Category Balanced} protocols on Twitter-17 (16-shot setting).}
    \label{fig8}
    \vskip -0.15in
\end{figure}

\begin{table*}[t]
\centering
\small
\vskip -0.15in
\caption{Accuracy comparison of MLLMs, SOTA few-shot models and SOTA fully-supervised models. We report the ICL performance of MLLMs under a 16-shot setting, as it includes sufficient demonstrations while not exceeding MLLMs' input limit.}
\vskip 0.05in
\resizebox{1\linewidth}{!}{
\begin{tabular}{clcccccccc}
\toprule
\multicolumn{2}{c}{\multirow{2}{*}{\textbf{Model \& Strategy}}} & \multirow{2}{*}{Support Set} & \multicolumn{3}{c}{Post-Level} & \multicolumn{3}{c}{Aspect-Level} & \multirow{2}{*}{Mean}    \\
\cmidrule(l){4-6} \cmidrule(l){7-9} 
                  &    &    & MVSA-S     & MVSA-M    & TumEmo    & Twitter-15  & Twitter-17 & MASAD &                          \\
\midrule
\multirow{10}{*}{\textbf{MLLM}} & Zero-Shot Paradigm & ---       & 38.6       & 56.5      & 46.6      & 60.7        & 54.7       & 73.0  & 55.0  \\
\cmidrule(l){2-10}
\multirow{10}{*}{IDEFICS} & ICL Random 16-shot & 1\% Training Set       & 49.9       & 60.4      & 52.1      & 61.1        & 57.5       & 77.2  & 59.7  \\
& ICL RICES 16-shot \cite{aaai2022rices} & 1\% Training Set  & 57.9 & 64.2 & 61.4 & 61.6 & 61.5 & 82.7 & 64.9        \\
& ICL SQPA 16-shot \cite{cvpr2024icl} & 1\% Training Set  & 53.6 & 62.2 & 64.6 & 55.8 & 50.4 & 74.2 & 60.1 \\
& ICL MMICES 16-shot \cite{iclrw2024mmices} & 1\% Training Set  & 59.2 & 65.3 & 60.5 & 61.6 & 56.6 & 80.5 & 63.9 \\
& \cellcolor[HTML]{E1FDFF}ICL Ours 16-shot  & \cellcolor[HTML]{E1FDFF}1\% Training Set    & \cellcolor[HTML]{E1FDFF}\textbf{66.5}      & \cellcolor[HTML]{E1FDFF}\textbf{67.7}    & \cellcolor[HTML]{E1FDFF}\textbf{63.4}      & \cellcolor[HTML]{E1FDFF}\textbf{67.0}    & \cellcolor[HTML]{E1FDFF}\textbf{62.0}       & \cellcolor[HTML]{E1FDFF}\textbf{88.2}  & \cellcolor[HTML]{E1FDFF}\textbf{69.1} \\
\cmidrule(l){2-10}
& ICL Random 16-shot  & Full Training Set      & 50.0       & 60.2      & 52.1      & 61.0        & 57.5     & 77.3  & 59.7                     \\
& ICL RICES 16-shot \cite{aaai2022rices} & Full Training Set      & 58.3       & 65.5      & 61.7      & 62.7        & 61.8       & 83.4  & 65.6 \\
& ICL SQPA 16-shot \cite{cvpr2024icl} & Full Training Set  & 55.8 & 63.2 & 59.7 & 57.3 & 52.1 & 75.1 & 60.5 \\
& ICL MMICES 16-shot \cite{iclrw2024mmices} & Full Training Set  & 63.0 & 64.6 & 62.3 & 59.2 & 54.5 & 83.5 & 64.5 \\
& \cellcolor[HTML]{E1FDFF}ICL Ours 16-shot  & \cellcolor[HTML]{E1FDFF}Full Training Set        & \cellcolor[HTML]{E1FDFF}\textbf{68.5}      & \cellcolor[HTML]{E1FDFF}\textbf{69.5}    & \cellcolor[HTML]{E1FDFF}\textbf{65.0}      & \cellcolor[HTML]{E1FDFF}\textbf{69.0}        & \cellcolor[HTML]{E1FDFF}\textbf{63.7}       & \cellcolor[HTML]{E1FDFF}\textbf{89.8}  & \cellcolor[HTML]{E1FDFF}\textbf{70.9} \\
\midrule
\multirow{10}{*}{\textbf{MLLM}} & Zero-Shot Paradigm & --- & 52.5       & 59.2      & 27.8      & 34.4        & 47.1       & 69.6  & 48.4  \\
\cmidrule(l){2-10}
\multirow{10}{*}{Open-Flamingo}  & ICL Random 16-shot  & 1\% Training Set      & 59.2       & 67.3      & 34.8      & 57.6        & 51.9       & 75.0  & 57.6    \\
& ICL RICES 16-shot \cite{aaai2022rices} & 1\% Training Set  & 61.5       & 67.9      & 42.0      & 61.2        & 55.8       & 84.7  & 62.2 \\
& ICL SQPA 16-shot \cite{cvpr2024icl} & 1\% Training Set  & 59.5 & 68.3 & 44.9 & 56.2 & 53.9 & 75.5 & 59.7 \\
& ICL MMICES 16-shot \cite{iclrw2024mmices} & 1\% Training Set  & 63.3 & 68.0 & 46.4 & 56.9 & 55.4 & 80.8 & 61.8\\
& \cellcolor[HTML]{E1FDFF}ICL Ours 16-shot  & \cellcolor[HTML]{E1FDFF}1\% Training Set  & \cellcolor[HTML]{E1FDFF}\textbf{65.2}       & \cellcolor[HTML]{E1FDFF}\textbf{68.5}      & \cellcolor[HTML]{E1FDFF}\textbf{52.7}      & \cellcolor[HTML]{E1FDFF}\textbf{68.8}        & \cellcolor[HTML]{E1FDFF}\textbf{61.4}       & \cellcolor[HTML]{E1FDFF}\textbf{87.6}  & \cellcolor[HTML]{E1FDFF}\textbf{67.4} \\ 
\cmidrule(l){2-10}
& ICL Random 16-shot  & Full Training Set      & 59.0       & 67.4      & 34.5      & 57.6        & 51.7       & 75.1  & 57.5    \\
& ICL RICES 16-shot \cite{aaai2022rices} & Full Training Set  & 62.2 & 68.0 & 44.3 & 61.5 & 56.4 & 85.5 & 63.0    \\
& ICL SQPA 16-shot \cite{cvpr2024icl} & Full Training Set  & 61.4 & 68.3 & 43.5 & 60.3 & 54.7 & 76.3 & 60.8 \\
& ICL MMICES 16-shot \cite{iclrw2024mmices} & Full Training Set  & 64.7 & 68.1 & 47.5 & 59.8 & 57.0 & 82.4 & 63.3 \\
& \cellcolor[HTML]{E1FDFF}ICL Ours 16-shot  & \cellcolor[HTML]{E1FDFF}Full Training Set        & \cellcolor[HTML]{E1FDFF}\textbf{66.3}      & \cellcolor[HTML]{E1FDFF}\textbf{68.7}    & \cellcolor[HTML]{E1FDFF}\textbf{54.1}      & \cellcolor[HTML]{E1FDFF}\textbf{70.4}        & \cellcolor[HTML]{E1FDFF}\textbf{62.6}       & \cellcolor[HTML]{E1FDFF}\textbf{89.3}  & \cellcolor[HTML]{E1FDFF}\textbf{68.6} \\
\midrule
\multicolumn{2}{l}{\textbf{SOTA Few-Shot Models}} & --- & \color[HTML]{757575}69.8 & \color[HTML]{757575}68.3 & \color[HTML]{757575}58.1 & \color[HTML]{757575}67.3 & \color[HTML]{757575}61.9 & \color[HTML]{757575}84.1 & \color[HTML]{757575}68.3 \\
\multicolumn{2}{l}{\textbf{SOTA Fully-Supervised Models}} & --- & \color[HTML]{757575}\textbf{79.2} & \color[HTML]{757575}\textbf{73.5} & \color[HTML]{757575}\textbf{66.7} & \color[HTML]{757575}\textbf{81.1} & \color[HTML]{757575}\textbf{76.4} & \color[HTML]{757575}\textbf{95.6} & \color[HTML]{757575}\textbf{78.8} \\
\bottomrule
\end{tabular}}
\label{tab4}
\vskip -0.15in
\end{table*}

\begin{table}[t]
\vskip -0.12in
\centering
\caption{Our final strategies for configuring ICL demonstrations on six MSA datasets. Due to the diversity of task focus and data distribution, the optimal strategies for the three factors of each dataset are different.}
\vskip 0.05in
\resizebox{1\linewidth}{!}{
\begin{tabular}{lccc}
\toprule
Dataset & Retrieval & Presentation & Distribution \\
\midrule
MVSA-S      & \textbf{WIT} $[\boldsymbol{\alpha}:\boldsymbol{\beta}=2:8]$             & Image, Text & Unlimited   \\
MVSA-M      & \textbf{WIT} $[\boldsymbol{\alpha}:\boldsymbol{\beta}=2:8]$             & Image, Text & Unlimited   \\
TumEmo      & \textbf{WIT} $[\boldsymbol{\alpha}:\boldsymbol{\beta}=2:8]$             & Image, Text & Unlimited   \\
Twitter-15  & \textbf{WITA} $[\boldsymbol{\alpha}:\boldsymbol{\beta}=7:3,(\boldsymbol{\alpha}+\boldsymbol{\beta}):\boldsymbol{\gamma}=2:8]$             & Image, Text & Category Balanced   \\
Twitter-17  &  \textbf{WITA} $[\boldsymbol{\alpha}:\boldsymbol{\beta}=7:3,(\boldsymbol{\alpha}+\boldsymbol{\beta}):\boldsymbol{\gamma}=2:8]$             & Image, Text & Category Balanced   \\
MASAD       &  \textbf{WITA} $[\boldsymbol{\alpha}:\boldsymbol{\beta}=7:3,(\boldsymbol{\alpha}+\boldsymbol{\beta}):\boldsymbol{\gamma}=2:8]$             & Image, Text & Unlimited   \\
\bottomrule
\end{tabular}}
\vskip -0.18in
\label{tab5}
\end{table}

On Twitter-17, most distribution protocols still adhere to the positive correlation between SLR-Overall and accuracy (\cref{fig7} (b-1)), except for the \textbf{Category Balanced} protocol. It substantiates SLR as a feasible indicator to reflect the efficacy of ICL sequences and drives us to explore the reasons behind this exception. In parallel to MVSA-S, the MLLM shows slightly lower precision than recall on positive and neutral samples (\cref{fig7} (b-2,3)). Whereas, on negative samples (\cref{fig7} (b-4)), recall remains remarkably lower than precision, which is intuitively illustrated in \cref{fig8}. This further validates that the predictive bias observed in MVSA-S is irrelevant to datasets or demonstrations, and might be intrinsic to the MLLM itself. Such bias could potentially originate from the curation of the pre-training data, where most negative instances are filtered out, impairing MLLM's discernment of negative samples. In contrast to MVSA-S, the rarest category on Twitter-17 is negative. Under this disparity, the \textbf{Category Balanced} protocol instead inclines the model to predict the negative category over the positive category, offsetting MLLM's predictive bias and attaining the peak performance across both categories.

To summarize, two conclusions have been deduced. Firstly, MLLM exhibits a predictive bias in sentiment analysis, with an inclination to avoid negative predictions. Secondly, in datasets with less prevalent negative samples, the \textbf{Category Balanced} protocol emerges as the optimal selection by mitigating MLLM's bias effectively. Otherwise, the \textbf{Unlimited} protocol is the preferable alternative.



\subsubsection{Overall Strategy}
The overall strategy is crafted by integrating three finely optimized factors, as presented in \cref{tab5}. \cref{tab4} compares our devised strategies and other models. Among the ICL baselines we compared, ``ICL Random'' randomly selects demonstrations from the support set. ``ICL RICES'' retrieves samples based solely on the similarity measured by the \textbf{WIT} strategy. ``ICL SQPA'' and ``ICL MMICES'' are originally designed for the Visual Question Answering task. The former assigns pseudo-labels to the samples in the support set, while the latter first retrieves images and then retrieves texts. In our case, pseudo-labels are assigned as random sentiment categories, and ``ICL MMICES'' is identical to the \textbf{I} strategy.

From \cref{tab4}, our devised strategies exhibit consistent and pronounced superiorities against other ICL baselines, extending to datasets and MLLMs unused in the exploration. By expanding the support set to the full training set, our approach enables both MLLMs to outperform the SOTA few-shot models. Notably, configuring demonstrations operates independently of annotations, ensuring that the annotated data required for each test sample remains constant. This renders MLLMs' ICL more efficient and accurate in MSA than few-shot models. However, compared to fully-supervised models, ICL still exhibits a performance gap, particularly noticeable in certain datasets. It suggests that the role of fine-tuning with specific data still cannot be easily replaced. Overall, \cref{tab4} demonstrates the effectiveness and generalizability of our devised ICL strategies, as well as the considerable potential of ICL for further development.

\section{Conclusion and Discussion}
In this paper, we seek to unleash the sentimental perception capability of MLLMs through ICL. As a pioneering effort that applies ICL to MSA, we conduct an in-depth investigation into three pivotal factors that influence the configuration of ICL demonstrations: similarity measurement, modality presentation, and sentiment distribution. For these factors, we optimize strategies by balancing the similarity components, weighing the richness of information against the complexity of inputs, and offsetting the sentimental predictive bias of MLLMs. Comprehensive experiments on six datasets and two MLLMs demonstrate the pronounced superiority of our strategies against other ICL baselines, validating their effectiveness and generalizability. Our findings confirm that MLLMs can perceive sentiment as competent as supervised models, paving the way for further research and exploration.


A primary limitation of this paper lies in the range of MLLMs investigated. The effectiveness of ICL heavily depends on the MLLMs themselves, however, some advanced MLLMs are beyond the research scope due to practical reasons. Despite our progress, multimodal ICL is still in its infancy compared to ICL on text modality. Other aspects of configuring ICL demonstrations also merit further investigation.

\section*{Acknowledgements}

This work is supported by the National Natural Science Foundation of China (Grant NO 62376266, 62406318 and 62441614), Key Laboratory of Ethnic Language Intelligent Analysis and Security Governance of MOE, Minzu University of China, Beijing, China.

\section*{Impact Statement}

This manuscript focuses on configuring ICL demonstrations to unleash MLLMs' sentimental perception capability. By investigating three factors, we aim to advance multimodal ICL in MLLMs and shed light on MLLMs' sentiment-related properties. During the investigation, we discover a sentimental predictive bias, later mitigating it on the inference level to facilitate fairness in MLLMs' sentiment prediction. We have not delved deeper into the source limitations of this bias, as it lies beyond the primary scope of our research.

However, systematic studies of these limitations can potentially contribute to both MLLMs and MSA. As a further discussion, we attribute these limitations to pretraining data rather than model architecture, which is validated by \citet{cvpr2024emovit}. By constructing emotion-related data, it enhances MLLMs' zero-shot performance on visual emotion recognition. This success has the potential to be replicated in solving the sentimental predictive bias.


\bibliography{main}

\begin{thebibliography}{64}
\providecommand{\natexlab}[1]{#1}
\providecommand{\url}[1]{\texttt{#1}}
\expandafter\ifx\csname urlstyle\endcsname\relax
  \providecommand{\doi}[1]{doi: #1}\else
  \providecommand{\doi}{doi: \begingroup \urlstyle{rm}\Url}\fi

\bibitem[Alayrac et~al.(2022)Alayrac, Donahue, Luc, Miech, Barr, and et. al.]{neurips2022flamingo}
Alayrac, J., Donahue, J., Luc, P., Miech, A., Barr, I., and et. al.
\newblock Flamingo: a visual language model for few-shot learning.
\newblock In \emph{{NeurIPS} 2022}, 2022.

\bibitem[Alimisis et~al.(2024)Alimisis, Mademlis, Radoglou-Grammatikis, Sarigiannidis, and Papadopoulos]{arxiv2024t2iapp}
Alimisis, P., Mademlis, I., Radoglou-Grammatikis, P., Sarigiannidis, P., and Papadopoulos, G.~T.
\newblock Advances in diffusion models for image data augmentation: A review of methods, models, evaluation metrics and future research directions, 2024.

\bibitem[Awadalla et~al.(2023)Awadalla, Gao, Gardner, Hessel, Hanafy, and et. al.]{arxiv2023openflamingo}
Awadalla, A., Gao, I., Gardner, J., Hessel, J., Hanafy, Y., and et. al.
\newblock Openflamingo: An open-source framework for training large autoregressive vision-language models, 2023.

\bibitem[Baldassini et~al.(2024)Baldassini, Shukor, Cord, Soulier, and Piwowarski]{cvprw2024rebut}
Baldassini, F.~B., Shukor, M., Cord, M., Soulier, L., and Piwowarski, B.
\newblock What makes multimodal in-context learning work?
\newblock In \emph{{CVPR} 2024 - Workshops}, pp.\  1539--1550, 2024.

\bibitem[Brown et~al.(2020)Brown, Mann, Ryder, Subbiah, Kaplan, and et. al.]{neurips2020gpt3}
Brown, T.~B., Mann, B., Ryder, N., Subbiah, M., Kaplan, J., and et. al.
\newblock Language models are few-shot learners.
\newblock In \emph{{NeurIPS} 2020}, 2020.

\bibitem[Cai et~al.(2023)Cai, Wang, Liang, Qin, Yang, Wong, and Xu]{emnlpf2023ner}
Cai, C., Wang, Q., Liang, B., Qin, B., Yang, M., Wong, K., and Xu, R.
\newblock In-context learning for few-shot multimodal named entity recognition.
\newblock In \emph{Findings of {EMNLP} 2023}, pp.\  2969--2979, 2023.

\bibitem[Chen et~al.(2023{\natexlab{a}})Chen, Yang, Huang, Wu, Wang, and Geng]{arxiv2023visualicl}
Chen, H., Yang, X., Huang, Y., Wu, Z., Wang, J., and Geng, X.
\newblock Manipulating the label space for in-context classification, 2023{\natexlab{a}}.

\bibitem[Chen et~al.(2024)Chen, Han, He, Buckley, Torr, Tresp, and Gu]{iclrw2024mmices}
Chen, S., Han, Z., He, B., Buckley, M., Torr, P., Tresp, V., and Gu, J.
\newblock Understanding and improving in-context learning on vision-language models.
\newblock In \emph{{ICLR} 2024 Workshop}, 2024.

\bibitem[Chen et~al.(2023{\natexlab{b}})Chen, Yuan, Tian, Geng, Li, Zhou, Metaxas, and Yang]{cvpr2023fdt}
Chen, Y., Yuan, J., Tian, Y., Geng, S., Li, X., Zhou, D., Metaxas, D.~N., and Yang, H.
\newblock Revisiting multimodal representation in contrastive learning: From patch and token embeddings to finite discrete tokens.
\newblock In \emph{{CVPR} 2023}, pp.\  15095--15104, 2023{\natexlab{b}}.

\bibitem[Dong et~al.(2024)Dong, Li, Dai, Zheng, Ma, Li, Xia, Xu, Wu, Chang, Sun, Li, and Sui]{arxiv2022iclsurvey}
Dong, Q., Li, L., Dai, D., Zheng, C., Ma, J., Li, R., Xia, H., Xu, J., Wu, Z., Chang, B., Sun, X., Li, L., and Sui, Z.
\newblock A survey on in-context learning, 2024.

\bibitem[Driess et~al.(2023)Driess, Xia, Sajjadi, Lynch, Chowdhery, and et. al.]{icml2023palme}
Driess, D., Xia, F., Sajjadi, M. S.~M., Lynch, C., Chowdhery, A., and et. al.
\newblock Palm-e: An embodied multimodal language model.
\newblock In \emph{{ICML} 2023}, volume 202 of \emph{Proceedings of Machine Learning Research}, pp.\  8469--8488, 2023.

\bibitem[Huang et~al.(2023)Huang, Dong, Wang, Hao, Singhal, and et. al.]{neurips2023kosmos}
Huang, S., Dong, L., Wang, W., Hao, Y., Singhal, S., and et. al.
\newblock Language is not all you need: Aligning perception with language models.
\newblock In \emph{{NeurIPS} 2023}, 2023.

\bibitem[Huang et~al.(2021)Huang, Du, Xue, Chen, Zhao, and Huang]{nips2021m>u}
Huang, Y., Du, C., Xue, Z., Chen, X., Zhao, H., and Huang, L.
\newblock What makes multi-modal learning better than single (provably).
\newblock In \emph{{NeurIPS} 2021}, pp.\  10944--10956, 2021.

\bibitem[Kosti et~al.(2017)Kosti, {\'{A}}lvarez, Recasens, and Lapedriza]{cvpr2017emotic}
Kosti, R., {\'{A}}lvarez, J.~M., Recasens, A., and Lapedriza, {\`{A}}.
\newblock Emotion recognition in context.
\newblock In \emph{{CVPR} 2017}, pp.\  1960--1968, 2017.

\bibitem[Lauren{\c{c}}on et~al.(2023)Lauren{\c{c}}on, Saulnier, Tronchon, Bekman, Singh, and et. al.]{nips2023obelics}
Lauren{\c{c}}on, H., Saulnier, L., Tronchon, L., Bekman, S., Singh, A., and et. al.
\newblock {OBELICS:} an open web-scale filtered dataset of interleaved image-text documents.
\newblock In \emph{{NeurIPS} 2023}, 2023.

\bibitem[Levy et~al.(2023)Levy, Bogin, and Berant]{acl2023coverls}
Levy, I., Bogin, B., and Berant, J.
\newblock Diverse demonstrations improve in-context compositional generalization.
\newblock In \emph{{ACL} 2023}, pp.\  1401--1422, 2023.

\bibitem[Li et~al.(2023)Li, Li, Savarese, and Hoi]{icml2023blip2}
Li, J., Li, D., Savarese, S., and Hoi, S. C.~H.
\newblock {BLIP-2:} bootstrapping language-image pre-training with frozen image encoders and large language models.
\newblock In \emph{{ICML} 2023}, volume 202 of \emph{Proceedings of Machine Learning Research}, pp.\  19730--19742, 2023.

\bibitem[Li et~al.(2024)Li, Peng, Chen, Gao, and Yang]{cvpr2024icl}
Li, L., Peng, J., Chen, H., Gao, C., and Yang, X.
\newblock How to configure good in-context sequence for visual question answering.
\newblock In \emph{{CVPR} 2024}, pp.\  26710--26720, 2024.

\bibitem[Li \& Qiu(2023)Li and Qiu]{emnlpf2023support}
Li, X. and Qiu, X.
\newblock Finding support examples for in-context learning.
\newblock In \emph{Findings of {EMNLP} 2023}, pp.\  6219--6235, 2023.

\bibitem[Li et~al.(2022)Li, Xu, Zhu, and Zhao]{naacl2022clmlf}
Li, Z., Xu, B., Zhu, C., and Zhao, T.
\newblock {CLMLF:} {A} contrastive learning and multi-layer fusion method for multimodal sentiment detection.
\newblock In \emph{Findings of {NAACL} 2022}, pp.\  2282--2294, 2022.

\bibitem[Lian et~al.(2024)Lian, Sun, Sun, Chen, Wen, Gu, Liu, and Tao]{if2024gpt4vemotion}
Lian, Z., Sun, L., Sun, H., Chen, K., Wen, Z., Gu, H., Liu, B., and Tao, J.
\newblock {GPT-4V} with emotion: {A} zero-shot benchmark for generalized emotion recognition.
\newblock \emph{Inf. Fusion}, 108:\penalty0 102367, 2024.

\bibitem[Ling et~al.(2022)Ling, Yu, and Xia]{acl2022vlp}
Ling, Y., Yu, J., and Xia, R.
\newblock Vision-language pre-training for multimodal aspect-based sentiment analysis.
\newblock In \emph{{ACL} 2022}, pp.\  2149--2159, 2022.

\bibitem[Liu et~al.(2022)Liu, Shen, Zhang, Dolan, Carin, and Chen]{deelio2022goodicl}
Liu, J., Shen, D., Zhang, Y., Dolan, B., Carin, L., and Chen, W.
\newblock What makes good in-context examples for gpt-3?
\newblock In Agirre, E., Apidianaki, M., and Vulic, I. (eds.), \emph{{DeeLIO@ACL} 2022}, pp.\  100--114, 2022.

\bibitem[Lu et~al.(2018)Lu, Neves, Carvalho, Zhang, and Ji]{acl2018t17}
Lu, D., Neves, L., Carvalho, V., Zhang, N., and Ji, H.
\newblock Visual attention model for name tagging in multimodal social media.
\newblock In \emph{{ACL} 2018}, pp.\  1990--1999, 2018.

\bibitem[Lu et~al.(2022)Lu, Bartolo, Moore, Riedel, and Stenetorp]{acl2022order}
Lu, Y., Bartolo, M., Moore, A., Riedel, S., and Stenetorp, P.
\newblock Fantastically ordered prompts and where to find them: Overcoming few-shot prompt order sensitivity.
\newblock In \emph{{ACL} 2022}, pp.\  8086--8098, 2022.

\bibitem[Lyu et~al.(2023)Lyu, Min, Beltagy, Zettlemoyer, and Hajishirzi]{acl2023zicl}
Lyu, X., Min, S., Beltagy, I., Zettlemoyer, L., and Hajishirzi, H.
\newblock {Z-ICL:} zero-shot in-context learning with pseudo-demonstrations.
\newblock In Rogers, A., Boyd{-}Graber, J.~L., and Okazaki, N. (eds.), \emph{{ACL} 2023}, pp.\  2304--2317, 2023.

\bibitem[Niu et~al.(2016)Niu, Zhu, Pang, and El{-}Saddik]{mmm2016mvsa}
Niu, T., Zhu, S., Pang, L., and El{-}Saddik, A.
\newblock Sentiment analysis on multi-view social data.
\newblock In \emph{{MMM} 2016}, volume 9517 of \emph{Lecture Notes in Computer Science}, pp.\  15--27, 2016.

\bibitem[Pan et~al.(2023)Pan, Gao, Chen, and Chen]{acl2023tr&tl}
Pan, J., Gao, T., Chen, H., and Chen, D.
\newblock What in-context learning "learns" in-context: Disentangling task recognition and task learning.
\newblock In \emph{Findings of {ACL} 2023}, pp.\  8298--8319, 2023.

\bibitem[Peng et~al.(2024)Peng, Li, Wang, Zhang, and Zhao]{aaai2024ebmm}
Peng, T., Li, Z., Wang, P., Zhang, L., and Zhao, H.
\newblock A novel energy based model mechanism for multi-modal aspect-based sentiment analysis.
\newblock In \emph{{AAAI} 2024}, pp.\  18869--18878, 2024.

\bibitem[Qin et~al.(2024)Qin, Chen, Fei, Chen, Li, and Che]{neurips2024icl}
Qin, L., Chen, Q., Fei, H., Chen, Z., Li, M., and Che, W.
\newblock What factors affect multi-modal in-context learning? an in-depth exploration.
\newblock In \emph{{NeurIPS} 2024}, 2024.

\bibitem[Radford et~al.(2021)Radford, Kim, Hallacy, Ramesh, Goh, and et. al.]{icml2021clip}
Radford, A., Kim, J.~W., Hallacy, C., Ramesh, A., Goh, G., and et. al.
\newblock Learning transferable visual models from natural language supervision.
\newblock In \emph{{ICML} 2021}, volume 139 of \emph{Proceedings of Machine Learning Research}, pp.\  8748--8763, 2021.

\bibitem[Rombach et~al.(2022)Rombach, Blattmann, Lorenz, Esser, and Ommer]{cvpr2022diffusion}
Rombach, R., Blattmann, A., Lorenz, D., Esser, P., and Ommer, B.
\newblock High-resolution image synthesis with latent diffusion models.
\newblock In \emph{{CVPR} 2022}, pp.\  10674--10685, 2022.

\bibitem[Shukor et~al.(2024)Shukor, Rame, Dancette, and Cord]{iclr2024beyond}
Shukor, M., Rame, A., Dancette, C., and Cord, M.
\newblock Beyond task performance: evaluating and reducing the flaws of large multimodal models with in-context-learning.
\newblock In \emph{{ICLR} 2024}, 2024.

\bibitem[Wang et~al.(2024)Wang, Chen, Han, Lin, Zhao, Liu, Zhai, Yuan, You, and Yang]{arxiv2024mllmsurvey}
Wang, Y., Chen, W., Han, X., Lin, X., Zhao, H., Liu, Y., Zhai, B., Yuan, J., You, Q., and Yang, H.
\newblock Exploring the reasoning abilities of multimodal large language models (mllms): {A} comprehensive survey on emerging trends in multimodal reasoning, 2024.

\bibitem[Wei et~al.(2023)Wei, Yuan, Yang, Shen, Li, Wang, and Chen]{acl2023mvcn}
Wei, Y., Yuan, S., Yang, R., Shen, L., Li, Z., Wang, L., and Chen, M.
\newblock Tackling modality heterogeneity with multi-view calibration network for multimodal sentiment detection.
\newblock In \emph{{ACL} 2023}, pp.\  5240--5252, 2023.

\bibitem[Wu et~al.(2024{\natexlab{a}})Wu, Yang, Zhou, and Ma]{mm2024drf}
Wu, D., Yang, D., Zhou, Y., and Ma, C.
\newblock Robust multimodal sentiment analysis of image-text pairs by distribution-based feature recovery and fusion.
\newblock In \emph{{ACM MM} 2024}, pp.\  5780--5789, 2024{\natexlab{a}}.

\bibitem[Wu et~al.(2024{\natexlab{b}})Wu, Yang, Zhou, and Ma]{mm2024pacl}
Wu, D., Yang, D., Zhou, Y., and Ma, C.
\newblock Bridging visual affective gap: Borrowing textual knowledge by learning from noisy image-text pairs.
\newblock In \emph{{ACM MM} 2024}, pp.\  602--611, 2024{\natexlab{b}}.

\bibitem[Wu et~al.(2023)Wu, Wang, Ye, and Kong]{acl2023saicl}
Wu, Z., Wang, Y., Ye, J., and Kong, L.
\newblock Self-adaptive in-context learning: An information compression perspective for in-context example selection and ordering.
\newblock In \emph{{ACL} 2023}, pp.\  1423--1436, 2023.

\bibitem[Xie et~al.(2024)Xie, Peng, Tseng, Chen, Hsu, Shuai, and Cheng]{cvpr2024emovit}
Xie, H., Peng, C., Tseng, Y., Chen, H., Hsu, C., Shuai, H., and Cheng, W.
\newblock Emovit: Revolutionizing emotion insights with visual instruction tuning.
\newblock In \emph{{CVPR} 2024}, pp.\  26586--26595, 2024.

\bibitem[Xu et~al.(2015)Xu, Ba, Kiros, Cho, Courville, Salakhutdinov, Zemel, and Bengio]{icml2015sat}
Xu, K., Ba, J., Kiros, R., Cho, K., Courville, A.~C., Salakhutdinov, R., Zemel, R.~S., and Bengio, Y.
\newblock Show, attend and tell: Neural image caption generation with visual attention.
\newblock In \emph{{ICML} 2015}, volume~37, pp.\  2048--2057, 2015.

\bibitem[Xu(2017)]{isi2017hsan}
Xu, N.
\newblock Analyzing multimodal public sentiment based on hierarchical semantic attentional network.
\newblock In \emph{{ISI} 2017}, pp.\  152--154, 2017.

\bibitem[Xu \& Mao(2017)Xu and Mao]{cikm2017multisentibank}
Xu, N. and Mao, W.
\newblock Multisentinet: {A} deep semantic network for multimodal sentiment analysis.
\newblock In \emph{{CIKM} 2017}, pp.\  2399--2402, 2017.

\bibitem[Xu et~al.(2018)Xu, Mao, and Chen]{sigir2018comn}
Xu, N., Mao, W., and Chen, G.
\newblock A co-memory network for multimodal sentiment analysis.
\newblock In \emph{{SIGIR} 2018}, pp.\  929--932, 2018.

\bibitem[Xu et~al.(2019)Xu, Mao, and Chen]{aaai2019mimn}
Xu, N., Mao, W., and Chen, G.
\newblock Multi-interactive memory network for aspect based multimodal sentiment analysis.
\newblock In \emph{{AAAI} 2019}, pp.\  371--378, 2019.

\bibitem[Yang et~al.(2024)Yang, Zhao, Wu, Wang, Zheng, Zhang, Che, and Qin]{arxiv2024msallmsurvey}
Yang, H., Zhao, Y., Wu, Y., Wang, S., Zheng, T., Zhang, H., Che, W., and Qin, B.
\newblock Large language models meet text-centric multimodal sentiment analysis: {A} survey, 2024.

\bibitem[Yang et~al.(2021{\natexlab{a}})Yang, Feng, Wang, and Zhang]{tmm2021tumemo}
Yang, X., Feng, S., Wang, D., and Zhang, Y.
\newblock Image-text multimodal emotion classification via multi-view attentional network.
\newblock \emph{{IEEE} Trans. Multim.}, 23:\penalty0 4014--4026, 2021{\natexlab{a}}.

\bibitem[Yang et~al.(2021{\natexlab{b}})Yang, Feng, Zhang, and Wang]{acl2021mgnns}
Yang, X., Feng, S., Zhang, Y., and Wang, D.
\newblock Multimodal sentiment detection based on multi-channel graph neural networks.
\newblock In \emph{{ACL/IJCNLP} 2021}, pp.\  328--339, 2021{\natexlab{b}}.

\bibitem[Yang et~al.(2023{\natexlab{a}})Yang, Feng, Wang, Zhang, and Poria]{mm2023multipoint}
Yang, X., Feng, S., Wang, D., Zhang, Y., and Poria, S.
\newblock Few-shot multimodal sentiment analysis based on multimodal probabilistic fusion prompts.
\newblock In \emph{{ACM MM} 2023}, pp.\  6045--6053, 2023{\natexlab{a}}.

\bibitem[Yang et~al.(2023{\natexlab{b}})Yang, Wu, Feng, Wang, Wang, Li, Sun, Zhang, Fu, and Poria]{arxiv2023mmbigbench}
Yang, X., Wu, W., Feng, S., Wang, M., Wang, D., Li, Y., Sun, Q., Zhang, Y., Fu, X., and Poria, S.
\newblock Mm-bigbench: Evaluating multimodal models on multimodal content comprehension tasks, 2023{\natexlab{b}}.

\bibitem[Yang et~al.(2023{\natexlab{c}})Yang, Wu, Yang, Chen, and Geng]{nips2023icicl}
Yang, X., Wu, Y., Yang, M., Chen, H., and Geng, X.
\newblock Exploring diverse in-context configurations for image captioning.
\newblock In \emph{{NeurIPS} 2023}, 2023{\natexlab{c}}.

\bibitem[Yang et~al.(2022)Yang, Gan, Wang, Hu, Lu, Liu, and Wang]{aaai2022rices}
Yang, Z., Gan, Z., Wang, J., Hu, X., Lu, Y., Liu, Z., and Wang, L.
\newblock An empirical study of {GPT-3} for few-shot knowledge-based {VQA}.
\newblock In \emph{{AAAI} 2022}, pp.\  3081--3089, 2022.

\bibitem[Yin et~al.(2023)Yin, Fu, Zhao, Li, Sun, Xu, and Chen]{arxiv2023surveyMLLM}
Yin, S., Fu, C., Zhao, S., Li, K., Sun, X., Xu, T., and Chen, E.
\newblock A survey on multimodal large language models, 2023.

\bibitem[Yu \& Jiang(2019)Yu and Jiang]{ijcai2019tombert}
Yu, J. and Jiang, J.
\newblock Adapting {BERT} for target-oriented multimodal sentiment classification.
\newblock In Kraus, S. (ed.), \emph{{IJCAI} 2019}, pp.\  5408--5414, 2019.

\bibitem[Yu et~al.(2022)Yu, Zhang, and Li]{mm2022up}
Yu, Y., Zhang, D., and Li, S.
\newblock Unified multi-modal pre-training for few-shot sentiment analysis with prompt-based learning.
\newblock In \emph{{ACM MM} 2022}, pp.\  189--198, 2022.

\bibitem[Yue et~al.(2019)Yue, Chen, Li, Zuo, and Yin]{kis2019survey2}
Yue, L., Chen, W., Li, X., Zuo, W., and Yin, M.
\newblock A survey of sentiment analysis in social media.
\newblock \emph{Knowl. Inf. Syst.}, 60\penalty0 (2):\penalty0 617--663, 2019.

\bibitem[Zadeh et~al.(2017)Zadeh, Chen, Poria, Cambria, and Morency]{emnlp2017tfn}
Zadeh, A., Chen, M., Poria, S., Cambria, E., and Morency, L.
\newblock Tensor fusion network for multimodal sentiment analysis.
\newblock In \emph{{EMNLP} 2017}, pp.\  1103--1114, 2017.

\bibitem[Zhang et~al.(2018{\natexlab{a}})Zhang, Wang, and Liu]{dmkd2018survey1}
Zhang, L., Wang, S., and Liu, B.
\newblock Deep learning for sentiment analysis: {A} survey.
\newblock \emph{WIREs Data Mining Knowl. Discov.}, 8\penalty0 (4), 2018{\natexlab{a}}.

\bibitem[Zhang et~al.(2018{\natexlab{b}})Zhang, Fu, Liu, and Huang]{aaai2018t15}
Zhang, Q., Fu, J., Liu, X., and Huang, X.
\newblock Adaptive co-attention network for named entity recognition in tweets.
\newblock In \emph{{AAAI} 2018}, pp.\  5674--5681, 2018{\natexlab{b}}.

\bibitem[Zhang et~al.(2023)Zhang, Pan, and Wang]{cvpr2023ler}
Zhang, S., Pan, Y., and Wang, J.~Z.
\newblock Learning emotion representations from verbal and nonverbal communication.
\newblock In \emph{{CVPR} 2023}, pp.\  18993--19004, 2023.

\bibitem[Zhang et~al.(2022)Zhang, Feng, and Tan]{emnlp2022r-icl}
Zhang, Y., Feng, S., and Tan, C.
\newblock Active example selection for in-context learning.
\newblock In \emph{{EMNLP} 2022}, pp.\  9134--9148, 2022.

\bibitem[Zhao et~al.(2022)Zhao, Yao, Yang, Jia, Ding, Chua, Schuller, and Keutzer]{tpami2022review}
Zhao, S., Yao, X., Yang, J., Jia, G., Ding, G., Chua, T., Schuller, B.~W., and Keutzer, K.
\newblock Affective image content analysis: Two decades review and new perspectives.
\newblock \emph{Trans. Pattern Anal. Mach. Intell.}, 44\penalty0 (10):\penalty0 6729--6751, 2022.

\bibitem[Zhou et~al.(2021)Zhou, Zhao, Huang, Hu, and He]{nc2021masad}
Zhou, J., Zhao, J., Huang, J.~X., Hu, Q.~V., and He, L.
\newblock {MASAD:} {A} large-scale dataset for multimodal aspect-based sentiment analysis.
\newblock \emph{Neurocomputing}, 455:\penalty0 47--58, 2021.

\bibitem[Zhou et~al.(2023)Zhou, Guo, Liu, Yu, Zhang, and Yuan]{acl2023aom}
Zhou, R., Guo, W., Liu, X., Yu, S., Zhang, Y., and Yuan, X.
\newblock Aom: Detecting aspect-oriented information for multimodal aspect-based sentiment analysis.
\newblock In \emph{Findings of {ACL} 2023}, pp.\  8184--8196, 2023.

\bibitem[Zhu et~al.(2024)Zhu, Guo, Feng, Huang, Feng, Wang, and Wang]{cc2024msareview}
Zhu, X., Guo, C., Feng, H., Huang, Y., Feng, Y., Wang, X., and Wang, R.
\newblock A review of key technologies for emotion analysis using multimodal information.
\newblock \emph{Cogn. Comput.}, 16\penalty0 (4):\penalty0 1504--1530, 2024.

\end{thebibliography}
\bibliographystyle{icml2025}

\newpage
\appendix
\onecolumn
\section{Textual Prompts}
The textual prompt $\mathcal{P}$ of multimodal ICL sequence $\mathcal{S}$ aims to provide the task description and the set of target categories. 

\subsection{Aspect-Level MSA}
$\mathcal{P}$ is ``A post contains an image, a text and an aspect. Identify the sentiment of the aspect in the post. The optional categories are [Positive, Neutral, and Negative]. Here are some examples''. 

\subsection{Post-Level MSA}
$\mathcal{P}$ is ``A post contains an image and a text. Classify the sentiment of the post into [Positive, Neutral, Negative]. Here are some examples''.

\subsection{Sensitivity of ICL to Prompt Variations}
In the investigation, we experiment with various textual prompts and find that they significantly impact zero-shot performance. However, their impact on ICL is minimal. Since this manuscript primarily focuses on how ICL configurations influence MLLMs' sentiment perception capabilities, we select a set of appropriate textual prompts and keep them fixed throughout the investigation. The performance of IDEFICS under different prompts is reported in \cref{prompt}. 

For post-level MSA:

\#1 Prompt: A post contains an image and a text. Classify the sentiment of the post into [Positive, Neutral, Negative].

\#2 Prompt: Please classify the sentiment of the image-text post into [Positive, Neutral, Negative].

\#3 Prompt: Here is a post containing an image and a text. The optional categories are [Positive, Neutral, Negative]. What is the overall sentiment of the post?

For aspect-level MSA:

\#1 Prompt: A post contains an image, a text and an aspect. Identify the sentiment of the aspect in the post. The optional categories are [Positive, Neutral, Negative].

\#2 Prompt: Please classify the sentiment of the aspect in image-text post into [Positive, Neutral, Negative].

\#3 Prompt: Here is a post containing an image, a text and an aspect. The optional categories are [Positive, Neutral, Negative]. What is the sentiment of the aspect in the post?

\begin{table*}[h]
\centering
\caption{Influence of prompt variations on accuracy of IDEFICS.}
\vskip 0.05in
\resizebox{1\linewidth}{!}{
\begin{tabular}{clccccccc}
\toprule
\multicolumn{2}{c}{\multirow{2}{*}{\textbf{Model \& Strategy}}} & \multirow{2}{*}{Support Set} & \multicolumn{3}{c}{MVSA-S} & \multicolumn{3}{c}{Twitter-15} \\
\cmidrule(l){4-6} \cmidrule(l){7-9} 
                  &    &    & \#1 Prompt     & \#2 Prompt    & \#3 Prompt    & \#1 Prompt  & \#2 Prompt  & \#3 Prompt      \\
\midrule
\multirow{2}{*}{\textbf{IDEFICS}} & Zero-Shot Paradigm & -       & 38.6       & 28.2      & 50.6      & 60.7        & 51.9       & 19.1 \\
& ICL Ours 16-shot  & 1\% Training Set  & 66.5 & 66.3 & 66.4 & 67.0 & 66.9 & 66.7  \\
\bottomrule
\end{tabular}}
\label{prompt}
\end{table*}

\section{Dataset Details}
The statistics of the adopted datasets are presented in \cref{tab6}. 

MVSA-S, MVSA-M \cite{mmm2016mvsa} are labeled on single modalities, where the sentiment categories include \textit{Positive, Neutral} and \textit{Negative}. The multimodal sentiment categories are obtained by majority voting following \citet{cikm2017multisentibank}. 

TumEmo \cite{tmm2021tumemo} is a weakly supervised dataset. Image-text posts are retrieved based on seven emotion keywords: \textit{Love, Happy, Calm, Bored, Sad, Angry, Fear}, and labeled accordingly.

Twitter-15 \cite{aaai2018t15}, Twitter-17 \cite{acl2018t17} are initially proposed for Multimodal Named Entity Recognition. Their named entities are later annotated by \citet{ijcai2019tombert} based on the sentiment polarities: \textit{Positive, Neutral, Negative}, and utilized for aspect-level MSA.

MASAD \cite{nc2021masad} extends textual aspects to visual aspects, and includes posts from more diverse domains. The aspects are labeled by sentiment polarities: \textit{Positive} and \textit{Negative}.

\begin{table}[h]
  \centering
  \caption{Statistics of datasets.}
  \vskip 0.1in
  \resizebox{0.5\linewidth}{!}{
  \begin{tabular}{cccccc}
    \toprule
    \multicolumn{2}{c}{\textbf{Dataset}} & \textbf{Train}& \textbf{Test} \\
    \midrule
    \multirow{3}{*}{\textbf{Post-Level}} & MVSA-S \cite{mmm2016mvsa} & 3608  & 452   \\
    & MVSA-M \cite{mmm2016mvsa} & 13618 & 1703  \\
    & TumEmo \cite{tmm2021tumemo} & 156217 & 19524  \\
    \midrule
    \multirow{3}{*}{\textbf{Aspect-Level}} & Twitter-15 \cite{aaai2018t15} & 3179  & 1037   \\
    & Twitter-15 \cite{acl2018t17} & 3562 & 1234  \\
    & MASAD \cite{nc2021masad} & 14868 & 4935  \\
    \bottomrule
  \end{tabular}}
  \label{tab6}
\end{table}

\section{Computational Overheads}

In the optimized configuration, presenting and distributing demonstrations do not introduce additional computational overhead. The extra costs originate from demonstration retrieval and the expanded input sequence for MLLMs. The former scales with the size of the support set, as each test sample needs to be compared against all support set samples, while the latter is inherent to ICL. We report the average time overhead (ms) of processing an image-text sample under two support set scales on a single NVIDIA GeForce RTX 4090 GPU.

\begin{table*}[h]
\centering
\caption{Comparison of time costs.}
\vskip 0.05in
\resizebox{0.7\linewidth}{!}{
\begin{tabular}{clcccc}
\toprule
\multicolumn{2}{c}{\multirow{2}{*}{\textbf{Model \& Strategy}}} & \multirow{2}{*}{Support Set} & \multicolumn{3}{c}{Time Overhead (ms)} \\
\cmidrule(l){4-6}
                  &    &    & Retrieval     & Inference    & Total  \\
\midrule
\multirow{7}{*}{\textbf{IDEFICS}} & Zero-Shot Paradigm & -       & 0       & 78.1      & 78.1   \\
\cmidrule(l){2-6}
& ICL Random 4-shot  & 136 / 1562 Samples  & 0 & 134.5 & 134.5 \\
& ICL Ours 4-shot  & 136 Samples  & 36.4 & 134.5 & 170.9 \\
& ICL Ours 4-shot  & 1562 Samples  & 64.2 & 134.5 & 198.7 \\
\cmidrule(l){2-6}
& ICL Random 16-shot  & 136 / 1562 Samples  & 0 & 346.1 & 346.1 \\
& ICL Ours 16-shot  & 136 Samples & 36.4 & 346.1 & 382.5 \\
& ICL Ours 16-shot  & 1562 Samples & 64.2 & 346.1 & 410.3 \\
\bottomrule
\end{tabular}}
\end{table*}

\section{Complete Results}
In the main paper, we simplify the reported results to emphasize the key findings. Here we present the complete results in numerical form.

\cref{tab2} reports the average accuracy across 4,8,16-shot demonstrations retrieved based on varying similarity measurements. \cref{tab7} is its complete version.

\cref{fig4} reports the average accuracy across 4,8,16-shot demonstrations retrieved based on the \textbf{WIT} and \textbf{WITA} strategies. \cref{tab8} is its complete version.

\cref{tab3} reports the average accuracy across 4,8,16-shot settings with the inputs composed of different modalities. \cref{tab9} is its complete version.

\cref{fig5} evaluates ICL’s “Task Learning” effect by progressively incorporating modalities into the inputs. \cref{tab10} is its complete version.

\begin{table*}
\centering
\caption{Complete results of \cref{tab2}.}
\vskip 0.1in
\resizebox{1\linewidth}{!}{
\begin{tabular}{lcccccccccccc}
\toprule
\multirow{3}{*}{Strategy} & \multicolumn{4}{c}{4-shot} & \multicolumn{4}{c}{8-shot} & \multicolumn{4}{c}{16-shot} \\
\cmidrule(l){2-5} \cmidrule(l){6-9} \cmidrule(l){10-13} 
 & \multicolumn{2}{c}{Post-Level} & \multicolumn{2}{c}{Aspect-Level} & \multicolumn{2}{c}{Post-Level} & \multicolumn{2}{c}{Aspect-Level} & \multicolumn{2}{c}{Post-Level} & \multicolumn{2}{c}{Aspect-Level} \\
\cmidrule(l){2-3} \cmidrule(l){4-5} \cmidrule(l){6-7} \cmidrule(l){8-9} \cmidrule(l){10-11} \cmidrule(l){12-13} 
        & MVSA-S           & MVSA-M          & Twitter-15      & Twitter-17   & MVSA-S           & MVSA-M          & Twitter-15      & Twitter-17 & MVSA-S           & MVSA-M          & Twitter-15      & Twitter-17  \\
\midrule
R      & 45.2&	59.7 &	56.1&	55.2&	50.8&	61.2&	58.8&	56.9&	51.4&	61.6&	57.3&	57.1 \\
\midrule
I      & 51.2&	64.3&	56.5&	56.8&	59.2&	65.2&	59.2&	56.7&	59.2&	65.3&	61.6&	56.6 \\
T      & 49.6&	64.6&	55.0&	55.2&	58.3&	67.2&	58.6&	56.8&	60.0&	66.8&	62.5&	58.9 \\
IT     & 49.5&	64.8&	60.5&	58.7&	57.7&	66.3&	62.2&	56.2&	60.1&	67.3&	61.6&	57.9 \\
\midrule
A      & -    &  -    &   60.2&	57.0&	-	&	 -   &   60.5&	57.6&	-	&	 -   &   63.4&	57.7 \\
IA     & -    &  -    &   57.1&	58.2&   -    &	-	&	59.5&	57.8&	-	&	 -   &   62.0&	57.9 \\
TA     & -    &  -    &   58.7&	56.6&	-	&	  -  &   61.6&	57.3&	-	&    -   & 	62.4&	58.0 \\
ITA    & -    &  -    &   59.3&	58.3&   -    &   -    &	61.6&	57.5&    -   &	-	&	62.2&	58.5 \\
\bottomrule
\end{tabular}}

\label{tab7}
\end{table*}

\begin{table*}
\centering
\caption{Complete results of \cref{fig4}.}
\vskip 0.1in
\resizebox{1\linewidth}{!}{
\begin{tabular}{c|cccccc|c|cccccc}
\toprule
\textbf{WIT} Strategy & \multicolumn{2}{c}{4-shot} & \multicolumn{2}{c}{8-shot} & \multicolumn{2}{c|}{16-shot} & \textbf{WITA} Strategy & \multicolumn{2}{c}{4-shot} & \multicolumn{2}{c}{8-shot} & \multicolumn{2}{c}{16-shot}\\
\cmidrule(l){2-3} \cmidrule(l){4-5} \cmidrule(l){6-7} \cmidrule(l){9-10} \cmidrule(l){11-12} \cmidrule(l){13-14} 
 $\boldsymbol{\alpha}:\boldsymbol{\beta}$       & MVSA-S           & MVSA-M          & MVSA-S           & MVSA-M   & MVSA-S           & MVSA-M       & $\boldsymbol{\alpha}:\boldsymbol{\beta}$; $(\boldsymbol{\alpha}+\boldsymbol{\beta}):\boldsymbol{\gamma}$  & Twitter-15      & Twitter-17 & Twitter-15      & Twitter-17          & Twitter-15      & Twitter-17  \\
\midrule
0:10 & 49.6&	64.6&	58.3&	67.2&	60.0&	66.8&  1:9;   8:2&56.7&	56.9&	59.0&	57.4&	62.8&	59.0
\\
1:9 & 51.7&	65.5&	58.5&	67.0&	61.4&	66.3&  1:9;   5:5&58.6&	56.4&	62.5&	57.5&	63.2&	58.5
\\
2:8 & 49.0&	65.3&	57.2&	67.1&	66.5&	67.7&  1:9;   2:8&58.9&	58.2&	60.8&	56.5&	63.0&	57.9
\\
3:7 & 50.2&	65.4&	57.2&	67.7&	64.9&	66.9&  3:7;   8:2&56.6&	55.8&	60.5&	55.9&	62.9&	59.1
\\
4:6 & 49.0&	64.9&	57.7&	67.2&	62.7&	67.2&  3:7;   5:5&59.7&	58.2&	61.7&	57.3&	62.8&	58.4
\\
5:5 & 49.5&	64.4&	57.5&	66.6&	60.1&	67.3&  3:7;   2:8&60.2&	58.6&	59.8&	57.4&	63.2&	57.8
\\
6:4 & 49.5&	64.4&	58.1&	67.2&	59.4&	66.7&  5:5;   8:2&57.8&	57.1&	62.5&	56.5&	64.0&	58.1
\\
7:3 & 51.2&	64.2&	57.7&	66.8&	59.2&	66.9&  5:5;   5:5&58.8&	57.9&	60.2&	56.8&	62.0&	58.2
\\
8:2 & 51.2&	64.6&	58.8&	67.0&	57.7&	66.7&  5:5;   2:8&59.8&	59.7&	60.2&	57.5&	63.7&	58.7
\\
9:1 & 50.8&	64.8&	57.9&	65.6&	57.2&	66.2&  7:3;   8:2&55.7&	57.2&	58.3&	58.2&	62.4&	58.6
\\
10:0 & 51.2&	64.3&	59.2&	65.2&	59.2&	65.3&  7:3;   5:5&58.4&	57.5&	59.5&	56.4&	62.1&	57.9
\\
- & -&	-&	-&	-&	-&	-&  7:3;   2:8&60.5&	58.9&	60.5&	59.3&	64.4&	58.7
\\
- & -&	-&	-&	-&	-&	-&  9:1;   8:2&55.0&	58.0&	59.7&	57.7&	63.0&	58.5
\\
- & -&	-&	-&	-&	-&	-&  9:1;   5:5&56.8&	58.3&	59.4&	57.1&	61.9&	58.8
\\
- & -&	-&	-&	-&	-&	-&  9:1;   2:8&60.5&	58.8&	60.6&	57.9&	63.8&	59.0
\\
\bottomrule
\end{tabular}}

\label{tab8}
\end{table*}

\begin{table*}
\centering
\caption{Complete results of \cref{tab3}.}
\vskip 0.1in
\resizebox{1\linewidth}{!}{
\begin{tabular}{lcccccccccccc}
\toprule
\multirow{3}{*}{Modality} & \multicolumn{4}{c}{4-shot} & \multicolumn{4}{c}{8-shot} & \multicolumn{4}{c}{16-shot} \\
\cmidrule(l){2-5} \cmidrule(l){6-9} \cmidrule(l){10-13} 
 & \multicolumn{2}{c}{Post-Level} & \multicolumn{2}{c}{Aspect-Level} & \multicolumn{2}{c}{Post-Level} & \multicolumn{2}{c}{Aspect-Level} & \multicolumn{2}{c}{Post-Level} & \multicolumn{2}{c}{Aspect-Level} \\
\cmidrule(l){2-3} \cmidrule(l){4-5} \cmidrule(l){6-7} \cmidrule(l){8-9} \cmidrule(l){10-11} \cmidrule(l){12-13} 
        & MVSA-S           & MVSA-M          & Twitter-15      & Twitter-17   & MVSA-S           & MVSA-M          & Twitter-15      & Twitter-17 & MVSA-S           & MVSA-M          & Twitter-15      & Twitter-17  \\
\midrule
I    &  33.5&	45.7&	55.2&	51.7&	55.4&	57.8&	58.5&	53.8&	66.3&	66.5&	58.4&	54.7
\\
C    &  26.2&	35.1&	53.3&	49.8&	47.2&	52.3&	56.8&	54.1&	58.3&	61.1&	58.5&	52.4
\\
I, C  &  48.8&	54.9&	52.3&	50.1&	60.1&	60.9&	53.4&	50.1&	61.9&	62.0&	57.1&	50.1
  \\
\midrule
T     & 29.7&	47.6&	60.8&	58.9&	48.6&	56.3&	61.5&	59.1&	60.8&	64.2&	62.9&	59.0
\\
G      &31.7&	44.0&	54.3&	54.6&	48.3&	57.8&	54.9&	55.4&	60.3&	64.9&	55.2&	55.7
\\
T, G   &   34.2&	47.5&	49.0&	54.1&	50.6&	61.0&	55.4&	55.0&	58.3&	65.7&	57.6&	55.4
\\
\midrule
I, T    &49.5&	64.8&	60.5&	58.7&	57.7&	66.3&	62.2&	56.2&	60.1&	67.3&	61.6&	57.9
\\
I, G    & 47.3&	54.2&	53.1&	51.8&	56.4&	58.9&	55.0&	53.4&	59.8&	63.3&	55.4&	56.2
\\
C, T    &32.2&	54.5&	61.5&	56.4&	55.7&	61.8&	63.9&	57.3&	61.2&	64.6&	62.4&	56.3
 \\
C, G    &33.7&	41.0&	51.6&	51.5&	49.9&	56.8&	52.5&	54.3&	59.0&	63.2&	55.2&	53.5
\\
\midrule
I, C, T  &47.7&	63.3&	56.5&	54.9&	57.9&	64.0&	61.0&	56.2&	60.3&	65.8&	63.5&	55.8
\\
I, T, G  &45.0&	58.6&	58.0&	53.9&	53.9&	62.4&	60.3&	55.2&	53.4&	64.8&	61.0&	55.2
\\
C, T, G  & 36.1&	50.5&	51.0&	53.0&	48.6&	59.9&	57.8&	55.4&	57.7&	64.2&	60.9&	54.0
\\
I, C, G  &41.0&	52.3&	52.8&	51.1&	54.1&	57.5&	51.1&	52.4&	55.9&	61.2&	54.7&	52.9
\\
I, C, T, G& 41.7&	59.4&	54.7&	52.4&	51.2&	61.4&	59.4&	53.6&	54.8&	63.5&	62.2&	53.4
\\
\bottomrule
\end{tabular}}

\label{tab9}
\end{table*}

\begin{table*}
\centering
\caption{Complete results of \cref{fig5}.}
\vskip 0.1in
\resizebox{0.6\linewidth}{!}{
\begin{tabular}{lcccccc}
\toprule
\multirow{2}{*}{Modality} & \multicolumn{2}{c}{4-shot} & \multicolumn{2}{c}{8-shot} & \multicolumn{2}{c}{16-shot} \\
\cmidrule(l){2-3} \cmidrule(l){4-5} \cmidrule(l){6-7} 
        & Twitter-15      & Twitter-17   & Twitter-15      & Twitter-17 & Twitter-15      & Twitter-17  \\
\midrule
T    &  33.7&	43.0&	39.6&	44.9&	52.7&	48.5
\\
+I (I, T)    &  30.4&	43.5&	36.7&	44.7&	48.2&	46.7
\\
+C (I, C, T)  &  30.3&	43.7&	33.4&	43.3&	49.2&	45.7
  \\
+G (I, C, T, G)  &  23.2&	39.7&	35.8&	44.4&	49.9&	45.8
  \\
\bottomrule
\end{tabular}}

\label{tab10}
\end{table*}

\end{document}